%% file: main.tex
\pdfoutput=1

\documentclass[11pt]{article}

\usepackage{EMNLP2023}

\usepackage{times}
\usepackage{latexsym}

\usepackage[T1]{fontenc}

\usepackage[utf8]{inputenc}

\usepackage{microtype}

\usepackage{inconsolata}

\usepackage{microtype}
\usepackage{graphicx}
\usepackage{booktabs} 

\usepackage{hyperref}

\usepackage{amsmath}
\usepackage{amssymb}
\usepackage{mathtools}
\usepackage{amsthm}

\usepackage[capitalize,noabbrev]{cleveref}

\usepackage{bm}
\usepackage{bbm}

\usepackage{makecell}
\usepackage{multirow}
\usepackage{booktabs}
\usepackage{enumitem}
\usepackage{xspace}
\usepackage{xcolor}
\usepackage{color, colortbl}

\usepackage{latexsym}
\usepackage{cprotect}
\usepackage{hhline}
\usepackage{array}
\usepackage{float}
\usepackage[group-separator={,},output-decimal-marker = {.}, group-minimum-digits={3}]{siunitx}  
\usepackage{colortbl}
\usepackage{booktabs} 
\usepackage{multirow}

\input{math_definition.tex}

\input{macros}

\newif\ifdraft
\draftfalse
\ifdraft
  \newcommand{\beer}[1]{{\color{cyan}Beer: #1}\xspace}
  \newcommand{\linting}[1]{{\color{brown}Linting: #1}\xspace}
  \newcommand{\xichen}[1]{{\color{olive}Xi: #1}\xspace}
  \newcommand{\asht}[1]{{\color{orange}Ashish: #1}\xspace}
  \newcommand{\julien}[1]{{\color{magenta}Julien: #1}\xspace}
  \newcommand{\michal}[1]{{\color{teal}Michal: #1}\xspace}
  \newcommand{\idan}[1]{{\color{blue}Idan: #1}\xspace}
  \newcommand{\radu}[1]{{\color{red}Radu: #1}\xspace}
\else
  \newcommand{\beer}[1]{}
  \newcommand{\linting}[1]{}
  \newcommand{\xichen}[1]{}
  \newcommand{\asht}[1]{}
  \newcommand{\julien}[1]{}
  \newcommand{\michal}[1]{}
  \newcommand{\idan}[1]{}
  \newcommand{\radu}[1]{}
\fi

\title{MaXM: Towards Multilingual Visual Question Answering}
\author{Soravit Changpinyo, Linting Xue, Michal Yarom, Ashish V. Thapliyal\\
{\bf Idan Szpektor}, {\bf Julien Amelot}, {\bf Xi Chen}, {\bf Radu Soricut} \\
Google Research\\
\url{https://github.com/google-research-datasets/maxm}}

\begin{document}
\maketitle

\input{abs}
\input{01_intro}
\input{02_related}
\input{03_approach}
\input{04_data}
\input{05_eval}
\input{06_conclusions}

{\normalsize{\mypar{Acknowledgments} We would like to thank Jialin Wu, Kenton Lee, Tomer Levinboim, Nan Ding, Sarah Laszlo, Doron Kukliansky, and Tania Bedrax-Weiss for their feedback and discussion. Word clouds are generated from \url{https://www.jasondavies.com/wordcloud/}.}}

{\small
\bibliography{main}
\bibliographystyle{acl_natbib}
}

\appendix
\input{supp}

\end{document}

%% file: math_definition.tex
\usepackage{amssymb}
\usepackage{amsmath,amsfonts}
\usepackage{amsopn}
\usepackage{bm} 
\usepackage{multirow}
\newlength\savewidth

\newcommand{\ProbOpr}[1]{\mathbb{#1}}

\newcommand{\expect}[2]{%
\ifthenelse{\equal{#2}{}}{\ProbOpr{E}_{#1}}
{\ifthenelse{\equal{#1}{}}{\ProbOpr{E}\left[#2\right]}{\ProbOpr{E}_{#1}\left[#2\right]}}} 
\newcommand{\var}[2]{%
\ifthenelse{\equal{#2}{}}{\ProbOpr{VAR}_{#1}}
{\ifthenelse{\equal{#1}{}}{\ProbOpr{VAR}\left[#2\right]}{\ProbOpr{VAR}_{#1}\left[#2\right]}}} 

\newcommand{\eat}[1]{}

%% file: macros.tex
\newcommand{\mypartop}[1]{\vspace{0mm}\noindent\textbf{#1}.}
\newcommand{\mypar}[1]{\vspace{0.5em}\noindent\textbf{#1}.}

\newcommand{\qsq}{$\mathrm{VQ}^{2}\!\mathrm{A}$\xspace}
\newcommand{\bqsq}{$\mathbf{\mathrm{VQ}^{2}\!\mathrm{A}}$\xspace}
\newcommand{\dtqsq}{$\mathrm{TransVQ}^{2}\!\mathrm{A}$\xspace}
\newcommand{\mvr}{$\mathrm{MAVERICS}$\xspace}
\newcommand{\xmodal}{$\mathrm{XM3600}$\xspace}
\newcommand{\maxm}{$\mathrm{MaXM}$\xspace}

\newcommand{\lang}{$\langle\mathrm{lang}\rangle$\xspace}

\newcommand{\vqaset}{VQA2.0\xspace}

\newcommand{\cocoqa}{COCOQA\xspace}

\newcommand{\qsqcc}{\qsq-CC3M\xspace}
\newcommand{\qsqcoco}{\qsq-COCO\xspace}

\newcommand{\correct}{\textit{Correct}\xspace}
\newcommand{\almost}{\textit{Almost Correct}\xspace}
\newcommand{\incorrect}{\textit{Incorrect}\xspace}

%% file: abs.tex
\begin{abstract}
Visual Question Answering (VQA) has been primarily studied through the lens of the English language. Yet, tackling VQA in other languages in the same manner would require a considerable amount of resources. In this paper, we propose scalable solutions to multilingual visual question answering (mVQA), on both data and modeling fronts. We first propose a translation-based framework to mVQA data generation that requires much less human annotation efforts than the conventional approach of directly collection questions and answers. Then, we apply our framework to the multilingual captions in the Crossmodal-3600 dataset and develop an efficient annotation protocol to create \maxm{}, a test-only VQA benchmark in 7 diverse languages. Finally, we develop a simple, lightweight, and effective approach as well as benchmark state-of-the-art English and multilingual VQA models. We hope that our benchmark encourages further research on mVQA.
\end{abstract}

%% file: 01_intro.tex

\begin{figure}[ht!]
\centering
\resizebox{\linewidth}{!}{%
\includegraphics{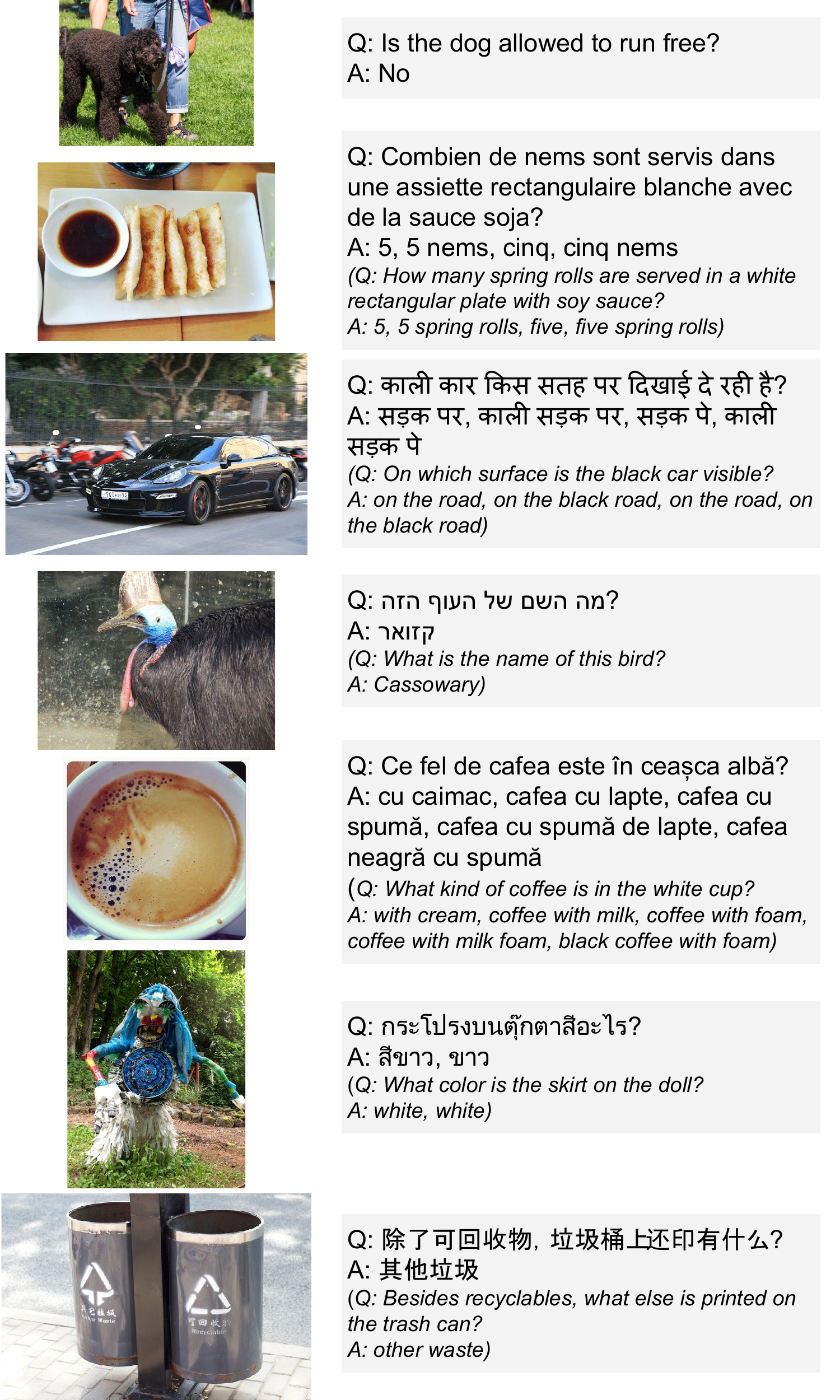}
}
\vspace{-3pt}
\caption{\textbf{Multilingual VQA Data in 7 languages.} The data is automatically generated from multilingual captions and then verified and adjusted by humans. From top to bottom: English (en), French (fr), Hindi (hi), Hebrew (iw), Romanian (ro), Thai (th), and Chinese (zh).}
\label{fig:intro}
\vspace{-15pt}
\end{figure}

\section{Introduction}
\label{sec:intro}

Visual Question Answering (VQA), the task of answering visual questions grounded in images, is key to human-machine interaction in the visual world. In particular, the natural language interface in VQA makes it easy for lay people to express their needs and benefit from its applications, including accessibility, education, and search.
Yet, VQA advances were mostly focused on English, therefore only applied to a privileged subset of human populations.

Arguably, the English language has dominated the field mainly because of the availability of English VQA benchmarks. These benchmarks are diverse, from general VQA~\cite{visual7w,tdiuc,krishnavisualgenome,vqa1,vqa2,changpinyo2022all}, robust VQA~\cite{vqacp}, compositional visual reasoning~\cite{gqa}, for the blind and the visually-impaired~\cite{vizwiz}, scene-text understanding~\cite{textvqa,stvqa}, to VQA that requires external, commonsense, or world knowledge~\cite{okvqa,vcr,aokvqa}. These benchmarks require considerable amount of resources to create, mostly by employing human annotators to laboriously collect and verify the questions and the answers for each image.

To extend VQA to all languages in the world, we must make data creation more automatic.
Building on recent work on automatic data creation for English VQA from captions~\cite{changpinyo2022all}, in this paper we propose a translation-based framework for multilingual visual question answering (mVQA) data creation. Our framework automates much of the task of generating questions and answers, thus providing a scalable path to mVQA.

We apply our framework to the generation of question-answer pairs from the multilingual captions of the recently-proposed Crossmodal-3600 dataset (\xmodal)~\cite{xm3600}. Combined with an efficient human annotation protocol, we construct \mvr-\xmodal (\maxm), a test benchmark for mVQA in 7 languages (see examples in Fig.~\ref{fig:intro}).

Finally, we use this novel benchmark to drive progress in mVQA modeling and measure where we stand. We leverage advances in image modeling and multilingual modeling: ViT~\cite{vit} and mT5~\cite{mt5} and propose a unified, extensible, open-ended mVQA model, called Simple MPT, which is competitive to state-of-the-art English VQA models that we adapt to apply in the mVQA setting (OFA~\cite{ofa} and BLIP2~\cite{blip2}).
Overall, there exists a large room for improvement.

In summary, our main contributions are
(i) a scalable translation-based framework for mVQA data generation based on captions (Sect.~\ref{sec:approach});
(ii) an efficient annotation protocol, deriving a novel test benchmark called \mvr-\xmodal (\maxm) in 7 diverse languages: English, French, Hindi, Hebrew, Romanian, Thai and Chinese (Sect.~\ref{sec:maxm});
(iii) simple and lightweight mVQA modeling (Sect.~\ref{ssec:model}, Sect.~\ref{apdx:mpt}) with strong performance;
(iv) benchmarking (adaptations of) the state-of-the-art VQA models on \maxm (Sect.~\ref{ssec:exp_res}).

%% file: 02_related.tex
\section{Related Work}
\label{sec:related}

\subsection{VQA and Multilingual Multimodal Benchmarks}
\label{ssec:related_data}
English has been the primary language in which vision-and-language researchers study the VQA task, driven by the availability of data and benchmarks~\cite{visual7w,tdiuc,krishnavisualgenome,vqa1,vqa2,vqacp,vizwiz,okvqa,textvqa,stvqa,advqa,avqa,changpinyo2022all}.
The only exception is xGQA~\cite{xgqa}, an extension of the English GQA dataset~\cite{gqa}. xGQA consists of human translations of the 12,578 English questions from 398 images in the balanced testdev split of GQA in 8 typologically diverse languages: English, German, Portuguese, Russian, Indonesian, Bengali, Korean, and Chinese. Besides the differences in the languages considered, our proposed approach to mVQA data creation complements xGQA (see Sect.~\ref{ssec:maxm_analysis}).

Beyond mVQA, training and evaluation data for multilingual multimodal models is limited. For a review of previous work, we refer the reader to the Image-Grounded
Language Understanding Evaluation (IGLUE) benchmark~\cite{iglue}, where xGQA is a part of.
In general, early attempts often focus on Chinese~\cite{cococn,vatex}, Japanese~\cite{stair,xtd} and several Indo-European languages (e.g., German, French, and Czech)~\cite{multi30kde,multi30kfr,multi30kcs}.
However, there is a recent effort toward a wider variety of both languages and tasks.
Examples include image retrieval \cite{xtd} (also Russian, Korean, Turkish), visual natural language inference \cite{iglue} (also Arabic), multilingual visual reasoning \cite{marvl} (also Indonesian, Swahili, Tamil, Turkish), and vision-and-language navigation \cite{rar} (also Hindi, Telugu).
Notably, Wikipedia Image Text (WIT)~\cite{wit} provides a large-scale image-text dataset in 108 languages, automatically collected form Wikipedia, and Crossmodal-3600 (\xmodal)~\cite{xm3600} provides human-curated test-only image captions in 36 languages.
Our work builds on top of \xmodal{}, and the 7 languages that we consider are typologically, genealogically, and geographically diverse. 

\subsection{VQA Data Creation}
\label{ssec:related_datagen}
Previous work on VQA data creation relies heavily on humans to create questions and answers~\cite{visual7w,krishnavisualgenome,vqa2,vizwiz,okvqa}. Some works attempt to automate this process. CLEVR~\cite{clevr} uses a template-based approach, but it is based on synthetic images for which ground-truth annotations are available. GQA~\cite{gqa} follows a similar approach but instead starts from Visual Genome scene graphs~\cite{krishnavisualgenome}, which themselves require large annotation efforts.

More relevant are works that rewrite image captions or video transcripts as question-answer pairs. \cocoqa~\cite{ren2015exploring} uses a template-based approach that can only generate questions with one-word answers. WeaQA~\cite{banerjee2021weaqa} improves upon this with semantic role labeling, paraphrasing, and backtranslation.
Recently, \citet{changpinyo2022all} and \citet{yang2021just} leverage T5~\cite{t5} fine-tuned on question answering datasets, generating large-scale VQA datasets for images and videos, respectively. Our approach to mVQA data creation leverages \qsq{}, the approach in ~\cite{changpinyo2022all} (Sect.~\ref{ssec:approach_bg}). To the best of our knowledge, besides xGQA, no other prior work on VQA data generation considered languages beyond English.

%% file: 03_approach.tex

\begin{figure*}[t]
\centering
\resizebox{.8\linewidth}{!}{%
\includegraphics{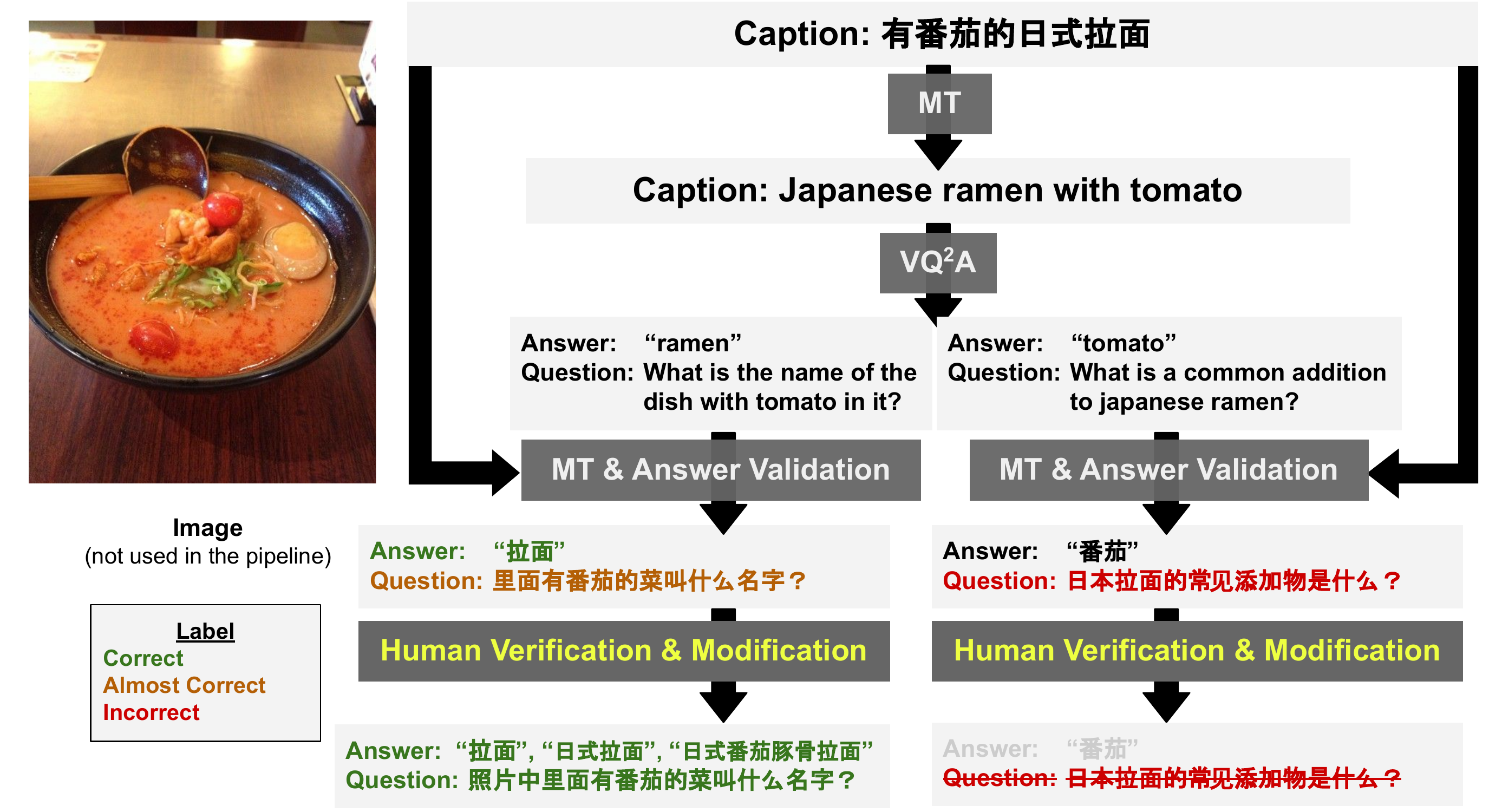}
}
\vspace{-5pt}
\caption{\textbf{Our approach to multilingual VQA data generation}, which is easy to scale, highly automatic and only requiring humans to modify ``Almost Correct'' questions or correct/expand answers (left) or filter out ``Incorrect'' questions(right). MT is short for automatic machine translation.}
\label{fig:create_data}
\vspace{-15pt}
\end{figure*}

\section{Multilingual VQA Data Creation}
\label{sec:approach}

Like in many other machine learning tasks, the main bottleneck to mVQA is obtaining high-quality labeled data. The most popular data collection framework to English VQA is to ask a set of human annotators to come up with visual questions, and another set of annotator to answer them (Sect.~\ref{ssec:related_datagen}). To scale VQA to all languages, we argue that mVQA data creation must significantly reduce its use of human annotation. To this end, we study the extension of an automatic English VQA data creation method called \textbf{V}isual \textbf{Q}uestion Generation with \textbf{Q}uestion \textbf{A}nswering validation, or \qsq~\cite{changpinyo2022all} for the purpose of mVQA data creation.

\subsection{Background: \qsq{}}
\label{ssec:approach_bg}

The \qsq{} approach leverages aligned image-text data sources that are available at scale~\cite{sbucap,cococap,cc3m,locnarr,cc12m,redcaps,laion400m} and beyond English~\cite{wit,wukong}. It rewrites a declarative image caption into multiple interrogative question-answer pairs via three steps: (i) \emph{Candidate Answer Extraction} extracts candidate answers based on syntactic and semantic analysis of an input caption, (ii) \emph{Question Generation} generates candidate questions for each candidate answer, (iii) \emph{Answer Validation} filters candidate questions that do not pass a consistency check that involves automatically answering each question from the caption and comparing this answer to the original extracted answer \cite{alberti-etal-2019-synthetic,honovich2021q2}.

Each step in \qsq{} is optimized for English; Step (i) uses English spaCy and both Step (ii) and Step (iii) leverage high-capacity English-pre-trained T5 models fine-tuned on English question answering datasets~\cite{squad,squad2,nq}.

\subsection{Translation-based \qsq{} (\dtqsq{})}
\label{ssec:approach_dtvq2a}
Inspired by \qsq{}, our goal is to generate mVQA data at scale, leveraging multilingual image captions. Multilingualizing each step in \qsq{} can be non-trivial and resource-intensive due to the heavy reliance of English tools, models, and data (Sect.~\ref{ssec:approach_bg}). To alleviate this, we propose a translation-based extension of \qsq{}.

Given an input caption $c$ in any language, and a target language \lang, we want to generate question-answer pairs in \lang. We propose Translation-based \qsq{} (\dtqsq{}), as follows:
\emph{\textbf{Step 1} Caption Translation}: Automatically translate a non-English caption $c$ to English $c_e$.
\emph{\textbf{Step 2} Apply \bqsq{}}: Generate a set of English question-answer pairs $\{q_e, a_e\}$ from $c_e$.
\emph{\textbf{Step 3} Question-Answer Translation}: Automatically translate all ($q_e, a_e$) pairs to \lang ($q, a$).
\emph{\textbf{Step 4} Validation}: Filter ($q, a$) pairs\footnote{Excluding answers to boolean questions.} in which $a$ does not appear in the original caption $c$, back-translating $a$ to $c$'s language if necessary. The upper part of Fig.~\ref{fig:create_data} exemplifies \dtqsq{} using a Chinese caption from Crossmodal-3600~\cite{xm3600}.

We highlight that the approach we have described so far is fully automatic and applicable to a huge set of languages that are supported by automatic translation.
We note that the final validation is important due errors that could pile up during translation steps. This is especially acute in Step 3, since translating answers is harder due to the lack of disambiguating context in the short answers.
We also note that \dtqsq{} can generate question/answer pairs in the target \lang from any caption. The output quality depends on the translation quality, e.g. the back-translation in step 4 from \lang to c's language. We use out-of-the-box translation tools in this work, and leave the exploration of better translation tailored for \dtqsq{} for future work.

In Sect.~\ref{sec:maxm} we employ human annotators to further clean and expand the generated data to create a high quality test benchmark.

\subsection{Direct Question Generation (DirectQG)} 
\label{ssec:approach_dqg}
One drawback of \dtqsq{} is the low coverage of particular types of answers, such as ``no''. This is because the captions generally do not indicate the absence of objects or properties (e.g., \emph{``There is no dog''}, \emph{``The dog is not white''}). 
To mitigate this bias, we train a multilingual question generator that takes in an answer and a caption in a target language and generates relevant questions in the same language. We use the model to generate questions for ``yes'', ``no'', or ``none'' as answers in each target language, as a complement to \dtqsq{}.

Concretely, we fine-tuned mT5-XXL~\cite{mt5} on large-scale translated COCO Captions~\cite{cococap} and its corresponding VQA data \qsqcoco{} ~\cite{changpinyo2022all}.
For validation, we used the subset of generated multilingual VQA data in Sect.~\ref{ssec:approach_dtvq2a}, with $\sim$300 golden examples for each language. The best checkpoint was selected based on ROUGE-L scores.

%% file: 04_data.tex

\begin{figure}[t]
\centering
\resizebox{\linewidth}{!}{%
\includegraphics{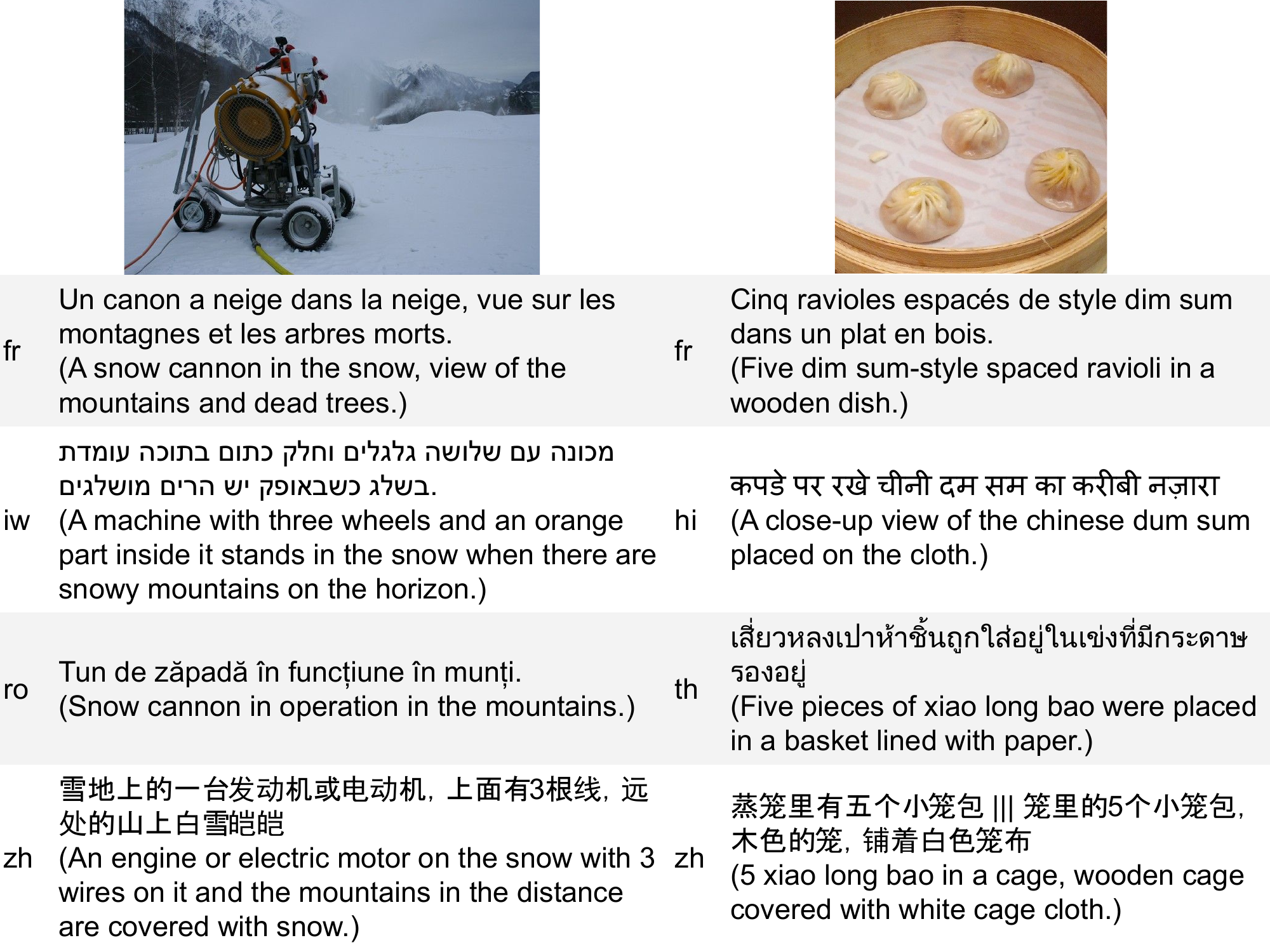}
}
\vspace{-3pt}
\caption{\textbf{The diversity of multilingual captions in \xmodal.} We show the captions (their English translations) from 4 languages for the images of a snow cannon (left) and xiao long bao (right).}
\label{fig:xm3600}
\vspace{-10pt}
\end{figure}

\begin{table*}[t]
\footnotesize
\begin{center}
\begin{tabular}{@{}l|c|c|c|c|c|c|c@{}}
\multicolumn{1}{c|}{} & en & fr & hi & iw & ro & th & zh \\ \hline
Captions & 7200 & 8562 & 8503 & 7200 & 7123 & 7200 & 7174  \\ \hline
English QAs & 373248 & 499900 & 520080 & 544268 & 516604 & 415180 & 524252 \\ \hline
Validated & 264930 & 343621 & 346948 & 375629 & 346887 & 286024 & 362304 \\
English QAs & (71.0\%) & (68.7\%) & (66.7\%) & (69.0\%) & (67.1\%) & (68.9\%) & (69.1\%) \\
\hline
Validated & 264724 & 122644 & 153465 & 128613 & 121221 & 95531 & 182095 \\ 
Multilingual QAs & (99.92\%) & (33.85\%) & (53.67\%) &(37.08\%) & (32.27\%) & (27.53\%) & (52.99\%) \\ \hline
\end{tabular}
\end{center}
\vspace{-12pt}
\caption{\textit{Number of Instances (\% of Previous Stage)} of automatically-generated question-answer (QA) pairs based on Crossmodal-3600 captions. Validated English pairs are w.r.t the QG-QA consistency filter. Validated multilingual pairs are w.r.t the caption-answer consistency filter.
}
\vspace{-15pt}
\label{tab:filtering}
\end{table*}

\begin{table*}[t]
\small
\begin{center}
\begin{tabular}{@{}c|c|c}
Label & \multicolumn{1}{c|}{Question} & \multicolumn{1}{c}{Answer} \\ \hline
\correct & Makes sense AND is relevant to the image. & Satisfies the question's intent wrt the image.\\ \hline
\almost & \multicolumn{2}{c}{Correct but its surface form can be improved (syntactic errors or awkward/uncommon usages.)} \\ \hline
\incorrect & \multicolumn{2}{c}{NOT Correct.} \\ \hline
\end{tabular}
\end{center}
\vspace{-10pt}
\caption{Definition of \correct, \almost and \incorrect labels for questions and answers in our annotation protocol.}
\vspace{-10pt}
\label{tab:anno}
\end{table*}

\begin{table*}[t]
\footnotesize
\begin{center}
\begin{tabular}{@{}l|c|c|c|c|c|c|c@{}}
 & \multicolumn{7}{c}{} \\  \cline{2-8}
\multicolumn{1}{c|}{} & en & fr & hi & iw & ro & th & zh \\ \hline
\# of questions evaluated & 377 & 389 & 400 & 365 & 440 & 401 & 391 \\ \hline
\% Correct & 62.6\% & 65.8\% & 66.5\% & 61.6\% & 59.1\% & 47.4\% & 51.9\% \\
\% Almost Correct & 17.5\% & 12.3\% & 9.0\% & 22.5\% & 9.3\% & 27.9\% & 25.1\% \\
\% Incorrect & 19.9\% & 21.9\% & 24.50\% & 15.9\% & 31.6\% & 24.7\% & 23.02\% \\
\hline \hline
\# of answers evaluated & 302 &	304 & 302 & 307 & 301 & 302 & 301 \\ \hline
\% Correct & 66.2\%	& 72.0\% & 77.8\% &	73.9\% & 76.2\% & 82.2\% & 81.9\% \\
\% Almost Correct & 26.5\% & 24.0\% & 17.9\% & 24.1\% & 16.9\% & 7.9\% & 9.2\% \\
\% Incorrect & 7.3\% & 3.9\% &	4.3\% &	2.0\% &	7.0\% &	9.9\% &	8.9\% \\
\hline
\end{tabular}
\end{center}
\vspace{-10pt}
\caption{\textbf{Human evaluation} of the generated questions and answers.
}
\vspace{-15pt}
\label{tab:correctness}
\end{table*}

\begin{table}[t]
\scriptsize
\begin{center}
\begin{tabular}{@{}l|r|r|r|r|r|r|r@{}}
\multicolumn{1}{c|}{Question} & \multicolumn{7}{c}{Percentage} \\  \cline{2-8}
\multicolumn{1}{c|}{Prefix} & \multicolumn{1}{c|}{en} & \multicolumn{1}{c|}{fr} & \multicolumn{1}{c|}{hi} & \multicolumn{1}{c|}{iw} & \multicolumn{1}{c|}{ro} & \multicolumn{1}{c|}{th} & \multicolumn{1}{c}{zh} \\ \hline
\emph{``is''} & 22.8 & 21.2 & 21.4 & 17.8 & 19.5 & 16.2 & 20.2 \\ 
\emph{``what is''} & 15.8 & 11.3 & 16.0 & 15.2 & 13.5 & 11.6 & 13.7 \\ 
\emph{``how many''} & 15.1 & 11.3 & 13.9 & 10.2 & 14.1 & 15.2 & 12.1 \\ 
\emph{``where''} & 6.7 & 9.2 & 7.5 & 8.6 & 6.6 & 9.3 & 5.9 \\ 
\emph{``what kind''} & 6.0 & 7.2 & 1.4 & 3.8 & 6.6 & 4.0 & 3.6 \\ 
\emph{``what are''} & 3.4 & 1.4 & 1.0 & 2.2 & 2.4 & 2.3 & 2.3 \\ 
\emph{``who''} & 3.4 & 3.1 & 9.2 & 2.5 & 1.5 & 2.6 & 2.3 \\ 
\emph{``are''} & 3.4 & 0.7 & 4.1 & 3.8 & 3.6 & 3.3 & 2.0 \\ 
\emph{``what color''} & 3.4 & 7.2 & 5.1 & 9.2 & 6.9 & 8.6 & 7.5 \\ 
\emph{``a''} & 3.0 & 3.1 & 1.4 & 3.5 & 2.1 & 3.0 & 2.9 \\ 
\emph{``what type''} & 2.7 & 2.4 & 0.3 & 0.3 & 2.1 & 1.0 & 4.6 \\ 
\emph{``what was''} & 1.0 & 0.3 & 0.0 & 0.6 & 1.2 & 3.3 & 0.7 \\ 
\emph{``do''} & 0.7 & 0.3 & 0.0 & 0.3 & 0.6 & 0.7 & 2.3 \\ 
\emph{``in''} & 0.7 & 1.7 & 0.3 & 0.3 & 2.4 & 1.7 & 2.0 \\ 
\emph{``besides''} & 0.3 & 1.0 & 4.4 & 3.2 & 1.2 & 0.3 & 2.3 \\ 
\emph{``does''} & 0.3 & 2.7 & 1.7 & 2.5 & 1.5 & 3.3 & 0.7 \\ 
other & 11.4 & 16.0 & 12.2 & 15.9 & 14.1 & 13.6 & 15.3 \\ 
\hline
\end{tabular}
\end{center}
\vspace{-10pt}
\caption{\textbf{The distribution of question types} in \maxm{} across languages. Approximated by their corresponding English question prefixes.}
\vspace{-15pt}
\label{tab:eqprefix_dist}
\end{table}

\section{MaXM: Multilingual VQA Benchmark}
\label{sec:maxm}

In this section, we leverage the approach we presented in Sect.~\ref{sec:approach} for creating a multilingual VQA test-only benchmark. We next describe our data sources, how candidate data was generated, human annotation protocol, and an analysis and a discussion of our benchmark. Following the naming convention in \cite{changpinyo2022all}, we call our benchmark MAVERICS-XM3600, or \maxm{} in short. We will release \maxm{} to foster research on mVQA.

\input{04_1_data_src}
\input{04_2_data_anno}
\input{04_3_data_analysis}

%% file: 04_1_data_src.tex

\subsection{Data Sources}
\label{ssec:maxm_src}
\mypartop{Language Selection} We chose 7 languages
that are 1) typologically diverse, 2) genealogically diverse, and 3) geographically diverse: English (en), French (fr), Hindi (hi), Hebrew (iw), Romanian (ro), Thai (th), and Chinese (zh).

\mypar{Image and Caption Selection} We chose a subset of the images in Crossmodal-3600 (\xmodal{})~\cite{xm3600}, in which high-quality multilingual image captions are available. For each language, 100 validation and test images of Open Images~\cite{openimages,oidv4} that were taken in the region(s) in which those languages were spoken were selected. 

Our image selection criteria cover a wide range of visual concepts in different cultural contexts, making the constructed VQA examples diverse and specific to the languages of the captions related to each image.
For example, in Fig.~\ref{fig:xm3600}, unlike French and Romanian speakers, Hebrew and Thai speakers are less likely to know what a snow cannon is. On the other hand, Thai and Chinese speakers are more likely to understand what xiao long bao is, whereas in French or Hindi it could be referred to as dim-sum ravioli or Chinese dim sum.

Another benefit of \xmodal{} is that the Open Images images are out-of-domain with respect to most widely-used VQA benchmarks~\cite{ren2015exploring,vqa1,visual7w,krishnavisualgenome,vqa2,vqacp,gqa,okvqa}, which are often based on MS-COCO images~\cite{coco}.

\subsection{Large-Scale mVQA Data Creation}
\label{ssec:maxm_approach}
We apply our approach described in Sect.~\ref{sec:approach} to the \xmodal{} captions to generate a large number of question-answer pairs for each language.

\mypartop{TransVQ$^2$A}
Table~\ref{tab:filtering} reports the number of question-answer pairs at different stages in our pipeline. Overall, we are able to generate a large number of question-answer pairs in all languages. We found that, across languages, approximately 30\% of (translated) English question-answer pairs are filtered out due to \qsq{} validation. 
In contrast, different percentages of translated answers across languages are filtered out based on the caption-answer consistency validation. A main reason for this is the quality of question-answer translation. For instance, 68\% of questions with ``alb" (masculine ``white'' in Romanian) are filtered out because they are not translated to the correct feminine form ``alba'' w.r.t the corresponding object in the question.

\mypar{DirectQG}
We augment the \dtqsq{} questions with additional candidate questions generated by \dtqsq{} (Sect.~\ref{ssec:approach_dqg}), using the \xmodal{} captions paired with ``yes'', ``no'', or ``none'' in their corresponding language as input. 

%% file: 04_2_data_anno.tex

\subsection{Human Annotation}
\label{ssec:maxm_human_anno}

We employed native speakers for each of the selected 7 languages to annotate and create our benchmark.
We designed an annotation protocol to balance efficiency and accuracy. In particular, we keep human-in-the-loop brief and only when an automated model straggles in a task, e.g., correcting translation artifacts, expanding answers, identifying sensitive questions. Furthermore, our protocol promotes quick discarding of examples, when the question does not make sense. We provide more details next and also in Appendix~\ref{apdx:anno}.

\mypar{Question and Answer Validation}
We define a 3-way rating system of \correct, \almost and \incorrect for both the questions and the answers (see Table~\ref{tab:anno}).
\correct questions are kept unchanged, \almost questions are manually rewritten, and \incorrect questions are discarded. Given \correct and \almost questions, an annotator rates the answer and corrects it in the cases of both \almost and \incorrect.

Table~\ref{tab:correctness} reports label distribution for questions and answers randomly-sampled from those generated by \dtqsq{}\footnote{For pairs generated by DirectQG, we did not perform exhaustive verification; we only asked the raters to annotate \correct and \almost questions related to ``no'' answers.}. Across languages, we observe at least 75\% \correct or \almost questions and, given these questions, at least 90\% \correct or \almost answers. This highlights the effectiveness of our approach.

\mypar{Answer Expansion and Standardization}
We split the generated questions into 4 categories: boolean, numeric, color, and others. We then asked the annotators to perform standardization of boolean, numeric, and color questions based on each language's guideline. For the rest of the questions, we tasked another set of at least 2 annotators per language to expand the answers to these questions with as many additionally correct (but not overly lengthy) answers as they can. 

\mypar{Additional Filtering}
Our raters performed another round of verification, filtering out examples with ``ambiguous'' and ``responsible-AI-sensitive'' questions and/or with inappropriate image content. The raters also labeled ``Collection'' questions that are likely to lead to long answers that are difficult to evaluate, such as for ``What is on the table?'' when there are multiple items, without filter them out.

%% file: 04_3_data_analysis.tex

\begin{figure}[t]
\centering
\resizebox{.9\linewidth}{!}{%
\includegraphics{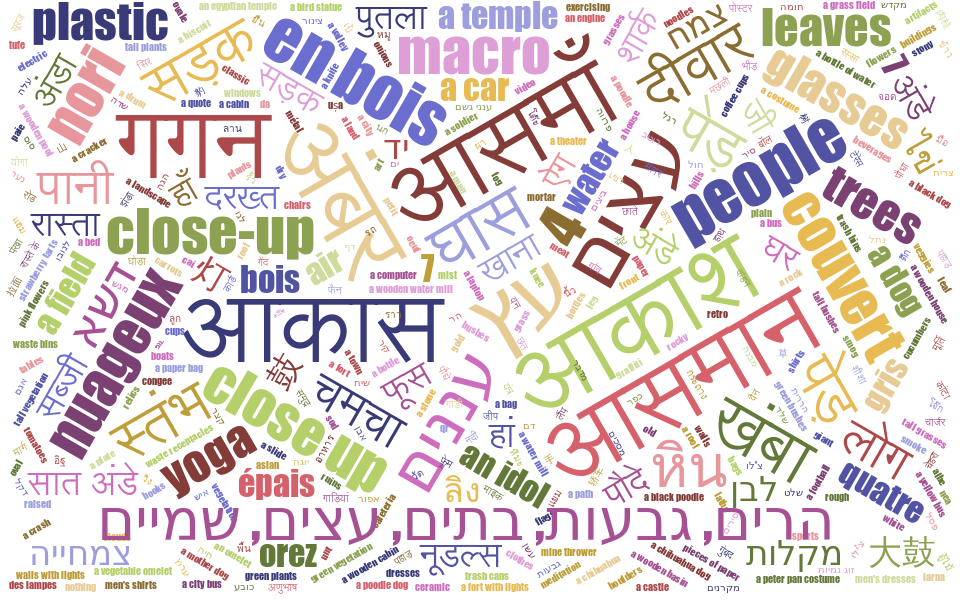}
}
\resizebox{.9\linewidth}{!}{%
\includegraphics{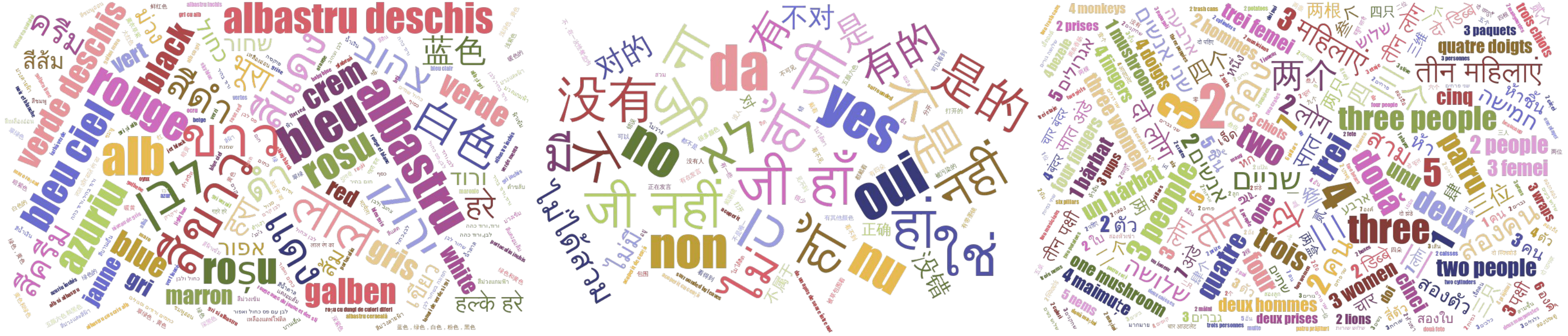}
}
\vspace{-10pt}
\caption{Top answer cloud is for \emph{``What''} questions (excluding \emph{``What color''}). Bottom answer clouds from left to right are for \emph{``What color''}, Boolean \emph{``Is/Are/Was/Were/Do/Does/Did''}, and \emph{``How many''} questions, respectively.}
\label{fig:answer_dist}
\vspace{-15pt}
\end{figure}

\subsection{Analysis and Discussion}
\label{ssec:maxm_analysis}

\mypartop{Size and Question Type and Answer Distributions}
\maxm{} v1 includes 2,142 questions in 7 languages: English (298), French (293), Hindi (294), Hebrew (315), Romanian (333), Thai (302), and Chinese (307).

Table~\ref{tab:eqprefix_dist} shows a breakdown of question types from \maxm{}. Since question prefixes in some languages are not indicative of question types (e.g., Thai does not always begin the "What" questions with the Thai "What"), we estimate a question's type using the prefix of its \textbf{English} version before translation. We observe diverse types and a high degree of linguistic variations.
Fig.~\ref{fig:answer_dist} presents word clouds for answers for selected question types: \emph{"What"}, \emph{"What color"}, Boolean, and \emph{"How many"}, to further illustrate the diverse answers within \maxm{} for each question type.

\mypartop{Comparison to xGQA}
In terms of settings, one difference is the languages of the answers; xGQA operates in the ``cross-lingual" setting where the input question can be non-English but the output answer is always English. While this simplifies the evaluation process, we argue that the ``multilingual" setting with non-English answers considered in \maxm{} is more practical. 

Another difference is the definition of the zero-shot setting; xGQA refers to unseen languages (not images) whereas our setting is more general, referring to both unseen images and languages. Finally, the type of translated data and how it is used for training are different; we only consider zero-shot setting and always use machine-translated questions for training, while xGQA considers both zero-shot and few-shot settings with human-translated questions involved only in the few-shot case.

In terms of the datasets, xGQA inherits the characteristics of GQA, whose questions are restricted in style (e.g., generated by a probabilistic template-based question engine) and in the skills required (e.g., reasoning-based with multi-step inference of object attributes and relationships)~\cite{gqa}. In contrast, \maxm{}'s questions are more general. Additionally, xGQA considers the same set of questions for all languages, whereas \maxm{} considers different sets of questions guided by the captions in each language.

%% file: 05_eval.tex

\input{05_res.tex}

\section{Evaluation}
\label{sec:eval}

\subsection{Evaluation Protocol}
\mypartop{Evaluation Metrics} We use Exact Match \emph{Accuracy} as the main evaluation measure for \maxm{}, following previous work on VQA~\cite{vqa1,vqa2,vizwiz}. We deem the answer as correct if it matches any of the ground-truth answers. To assess the degree of strictness of this measure, we also consider \emph{soft} text similarity metrics \emph{CIDEr}~\cite{cider} and \emph{ROUGE-L}~\cite{rouge} in our experiments, where we treat each of the ground-truth answers equally as one of the references (as if each of them was answered by an annotator).

\mypar{Training Data} MaXM is a test-only benchmark; it cannot be used for training. We designate \vqaset{}~\cite{vqa2} and its translations as the default training data source for our benchmark, due to its popularity and quality, similarly to the use of COCO-Captions~\cite{cococap} for the nocaps benchmark~\cite{nocaps} in the image captioning task. Nevertheless, we allow free use of existing VQA resources for training as long as the corresponding training images do not overlap with MaXM images. In our experiments, we also consider \qsqcoco{} and \qsqcc{}~\cite{changpinyo2022all} to assess the effect of text domain gap.

\subsection{Models for Multilingual VQA}
\label{ssec:model}

Inspired by approaches to multilingual NLP research, we consider two main families of models for mVQA that adapt existing source English VQA datasets to target languages: \emph{Translate-Test} and \emph{Translate-Train}. Translate-Test leaves the training data and the model as-is, but translates the test VQA data to the the source language English, apply the model, and then translate it back to the target language. On the other hand, \emph{Translate-Train} translates the English VQA data to a target language, trains a model on this pseudo VQA data (i.e., their translations), and directly apply the trained model to the test data.

\mypar{Translate-Test} We consider two open-source state-of-the-art VQA models: \textbf{OFA-Large}~\cite{ofa} and BLIP2~\cite{blip2}. Neither of them are designed for mVQA.

\mypar{Translate-Train} We include the results from the state-of-the-art multilingual vision-and-language model \textbf{PaLI-17B}~\cite{pali2}, which pretrains on diverse VQA datasets in 35 languages~\cite{xm3600} among other datasets, and then finetune on \vqaset{} in 13 languages: en, bn, de, fr, hi, id, iw, ko, pt, ro, ru, th, zh. Further, we implement a lightweight version of PaLI, called Simple Multi-Language Prompted Training, \textbf{Simple MPT}, with a much smaller model and without vision-and-language pre-training. \textbf{Simple MPT} is trained on the data in 13 languages in a multi-task fashion. Details can be found in Appendix~\ref{apdx:mpt}.

\subsection{Results}
\label{ssec:exp_res}

\mypartop{Main Results} Table~\ref{tab:benchmarking} benchmarks our proposed Simple MPT and state-of-the-art VQA models on \maxm{}. We observe that PaLI-17B performs best on all languages. This can be attributed to both the fact that PaLI is the strongest English VQA model and the fact that it was designed to be multilingual, leveraging pre-training image-text corpus in 105 languages. This result suggests it can be beneficial to design and develop multilingual VQA models from day one.

Surprisingly, our proposed Simple MPT model is a strong baseline even though it is much smaller than PaLI and does not leverage multilingual pre-training data. While its English performance is on par with OFA and much worse than BLIP2, its multilingual performance excels, outperforming OFA in all languages and underperforms BLIP2 only for Hindi and Hebrew.

Overall, our result suggests that \emph{Translate-Train} may be a superior approach to mVQA to \emph{Translate-Test}. We note, however, that in our early experiments, we find that Translate-Train is inferior to Translate-Test as an adaptation approach for \emph{English} VQA models. For instance, the answer of finetuned BLIP2 to the French question \emph{``Outre les fleurs roses, quelle autre couleur y avait-il dans le jardin?''} (\emph{``Besides pink flowers, what other color was there in the garden?”}) is \emph{``pink''} while the correct answer is \emph{``blanc''} (\emph{``white''}) — wrong both in terms of language and semantics. It is not immediately obvious how to adapt English VQA models with, for example, vocab and tokenizers that overfit the English language. This again suggests that the design of these multimodal models would benefit from having multilinguality in mind from the start.

\mypar{Single-Language vs. Multi-Language Training, Different Training Datasets} In Table~\ref{tab:training}, our Simple MPT model performs similarly or better than each of the Single-Language baselines. This suggests that modern models are capable of learning from related languages. We also find that translated COCO is overall the best training data source. We attribute this to (i) the fact that \qsq{} was used to generate \qsqcoco{}, and (ii) \qsqcoco{} is generally more robust in the cross-dataset setting \cite{changpinyo2022all}. However, \qsqcc{} is unable to outperform \vqaset{} despite (i); applying \qsq{} to the noisy alt-texts in CC3M~\cite{cc3m} is prone to errors that would only be exacerbated by automatic MT.

\mypar{Less Strict Metrics} In Table~\ref{tab:metrics} We observe generally consistent results when using CIDEr and ROUGE-L instead of the stricter Accuracy, except for Thai and Chinese, where the gaps in Accuracy are small to begin with.

\mypar{No Adaptation via Translate-Test} Can existing English VQA models \emph{work} out of the box? In Table~\ref{tab:noadapt}, we find that the answer is no. Expectedly, the models perform well on French, which is closer to English than other languages are. 

\mypar{Simple MPT on xGQA} Can our modeling approach be extended to the cross-lingual setting in xGQA~\cite{xgqa}? We report this result in Appendix~\ref{apdx:add_res}.

%% file: 05_res.tex

\begin{table*}[ht]
\footnotesize
\begin{center}
\begin{tabular}{c|cc|c|c|c|c|c|c|c@{}}
\multicolumn{3}{c|}{Model} & \multicolumn{7}{c}{Language} \\  \cline{4-10}
\multicolumn{3}{c|}{} & en & fr & hi & iw & ro & th & zh \\ \hline
\multirow{2}{*}{\emph{Translate-Test}} & OFA~\cite{ofa} & [470M] & 35.6 & 13.3 & 41.5 & 31.7 & 26.4 & 29.5 & 17.6 \\ 
& BLIP2~\cite{blip2} & [11B] & \textit{48.7} & 20.1 & \textit{63.3} & \textit{49.2} & 39.0 & 49.0 & 28.0 \\ \hline
\multirow{2}{*}{\emph{Translate-Train}} & Simple MPT (ours) & [1.5B] & 36.6 & \textit{36.2} & 55.1 & 40.6 & \textit{42.3} & \textit{50.0} & \textit{30.3} \\ 
& PaLI~\cite{pali2} & [17B] & \textbf{56.4} & \textbf{46.4} & \textbf{67.3} & \textbf{60.0} & \textbf{57.4} & \textbf{65.6} & \textbf{46.9} \\ \hline \hline
\end{tabular}
\end{center}
\vspace{-5pt}
\caption{\textbf{Benchmarking VQA models on \maxm{}}. Accuracy (\%) of OFA-Large (OFA), BLIP2, our lightweight Simple MPT, and PaLI-17B (PaLI), with approximate parameter count in brackets. All are finetuned on \vqaset{}, English-only for \emph{Translate-Train} and 13 languages for {Translate-Test}. Best results are \textbf{bold}. Second best \textit{italized}.}
\vspace{-5pt}
\label{tab:benchmarking}
\end{table*}

\begin{table*}[ht]
\footnotesize
\begin{center}
\begin{tabular}{c|c|c|c|c|c|c|c|c|c@{}}
\multicolumn{2}{c|}{Model} & \multicolumn{1}{c|}{Training} & \multicolumn{7}{c}{Language} \\  \cline{4-10}
\multicolumn{2}{c|}{} & \multicolumn{1}{c|}{Dataset} & en & fr & hi & iw & ro & th & zh \\ \hline
\multirow{4}{*}{\emph{Translate-Train}} & Single-Language & \vqaset{} & 37.6 & 33.8 & 53.7 & 35.6 & 36.0 & 50.0 & 29.0 \\ \cline{2-10}
& Simple MPT & \vqaset{} & 36.6 & 36.2 & 55.1 & 40.6 & 42.3 & 50.0 & 30.3 \\
& Simple MPT & \qsqcoco{} & \textbf{48.0} & \textbf{43.3} & \textbf{56.8} & \textbf{42.2} & \textbf{45.6} & \textbf{52.3} & \textbf{34.5} \\
& Simple MPT & \qsqcc{} & 38.3 & 34.1 & 45.6 & 34.9 & 36.9 & 45.0 & 29.0 \\ \hline
\end{tabular}
\end{center}
\vspace{-5pt}
\caption{\textbf{Effect of Training Data Sources}. Accuracy of Single-Language baselines (MPT architecture) and Accuracy (\%) of MPT models trained on different training datasets.}
\vspace{-5pt}
\label{tab:training}
\end{table*}

\begin{table*}[ht]
\footnotesize
\begin{center}
\begin{tabular}{c|c|c|c|c|c|c|c|c|c@{}}
\multicolumn{1}{c|}{Metric} & \multicolumn{2}{c|}{Model} & \multicolumn{7}{c}{Language} \\ \cline{4-10}
 & \multicolumn{2}{c|}{} & en & fr & hi & iw & ro & th & zh \\ \hline
\multirow{3}{*}{Accuracy} & \emph{Translate-Train} & Simple MPT & 36.6 & \textbf{36.2} & 55.1 & 40.6 & \textbf{42.3} & \textbf{50.0} & \textbf{30.3} \\ \cline{2-10}
& \multirow{2}{*}{\emph{Translate-Test}} & OFA & 35.6 & 13.3 & 41.5 & 31.7 & 26.4 & 29.5 & 17.6 \\ 
& & BLIP2 & \textbf{48.7} & 20.1 & \textbf{63.3} & \textbf{49.2} & 39.0 & 49.0 & 28.0 \\
\hline
\multirow{3}{*}{CIDEr} & \emph{Translate-Train} & Simple MPT & 91.5 & \textbf{102.0} & 62.0 & 78.6 & \textbf{89.6} & 86.7 & \textbf{68.4} \\ \cline{2-10}
& \multirow{2}{*}{\emph{Translate-Test}} & OFA & 88.3 & 49.4 & 66.3 & 78.6 & 60.4 & 70.2 & 51.1 \\ 
& & BLIP2 & \textbf{121.8} & 67.6 & \textbf{88.1} & \textbf{93.8} & 79.1 & \textbf{91.5} & 65.7 \\ \hline
\multirow{3}{*}{ROUGE-L} & \emph{Translate-Train} & Simple MPT & 45.0 & \textbf{47.9} & 57.9 & 42.8 & \textbf{49.4} & \textbf{57.7} & 37.7 \\ \cline{2-10}
& \multirow{2}{*}{\emph{Translate-Test}} & OFA & 47.5 & 27.2 & 52.0 & 44.8 & 38.3 & 44.0 & 31.4 \\ 
& & BLIP2 & \textbf{62.6} & 32.6 & \textbf{67.2} & \textbf{52.6} & 47.4 & 53.6 & \textbf{39.9} \\ \hline
\hline
\end{tabular}
\end{center}
\vspace{-5pt}
\caption{\textbf{Results on Soft Metrics}. Accuracy (\%), CIDEr ($\times$ 100), and ROUGE-L ($\times$ 100) of Simple MPT, OFA, and BLIP2.  All are finetuned on \vqaset{}. Best results are \textbf{bold}.}
\vspace{-5pt}
\label{tab:metrics}
\end{table*}

\begin{table*}[ht]
\footnotesize
\begin{center}
\begin{tabular}{c|c|c|c|c|c|c|c|c|c@{}}
\multicolumn{1}{c|}{Metric} & \multicolumn{2}{c|}{Model} & \multicolumn{7}{c}{Language} \\ \cline{4-10}
 & \multicolumn{2}{c|}{} & en & fr & hi & iw & ro & th & zh \\ \hline
\multirow{4}{*}{Accuracy} & \multirow{2}{*}{\emph{Translate-Test}} & OFA & 35.6 & 13.3 & 41.5 & 31.7 & 26.4 & 29.5 & 17.6 \\ 
& & BLIP2 & 48.7 & 20.1 & 63.3 & 49.2 & 39.0 & 49.0 & 28.0 \\ \cline{2-10}
& \multirow{2}{*}{No adapt} & OFA & 35.6 & 0.3 & 0.3 & 0.3 & 4.8 & 0.0 & 0.0 \\ 
& & BLIP2 & 48.7 & 9.2 & 0.3 & 1.9 & 3.9 & 7.0 & 2.3  \\ \hline
\multirow{4}{*}{CIDEr} & \multirow{2}{*}{\emph{Translate-Test}} & OFA & 88.3 & 49.4 & 66.3 & 78.6 & 60.4 & 70.2 & 51.1 \\ 
& & BLIP2 & 121.8 & 67.6 & 88.1 & 93.8 & 79.1 & 91.5 & 65.7 \\ \cline{2-10}
& \multirow{2}{*}{No adapt} & OFA & 88.3 & 4.8 & 3.7 & 4.8 & 12.6 & 7.8 & 3.6 \\ 
& & BLIP2 & 121.8 & 28.0 & 5.1 & 4.0 & 12.1 & 16.0 & 8.2  \\   \hline
\multirow{4}{*}{ROUGE-L} & \multirow{2}{*}{\emph{Translate-Test}} & OFA & 47.5 & 27.2 & 52.0 & 44.8 & 38.3 & 44.0 & 31.4 \\ 
& & BLIP2 & 62.6 & 32.6 & 67.2 & 52.6 & 47.4 & 53.6 & 39.9 \\ \cline{2-10}
& \multirow{2}{*}{No adapt} & OFA & 47.5 & 2.6 & 4.4 & 4.8 & 5.9 & 7.3 & 6.6 \\ 
& & BLIP2 & 62.6 & 12.5 & 6.2 & 2.9 & 5.4 & 10.8 & 7.3  \\ 
\hline
\end{tabular}
\end{center}
\vspace{-5pt}
\caption{\textbf{Results without adaptation for Translate-Test}. Accuracy (\%), CIDEr ($\times$ 100), and ROUGE-L ($\times$ 100) of OFA and BLIP2 with Translate-Test and without (No adapt).}
\vspace{-5pt}
\label{tab:noadapt}
\end{table*}

%% file: 06_conclusions.tex
\section{Conclusions}
\label{sec:conclusions}
We take initial steps toward multilingual VQA by proposing scalable solutions on both data creation and modeling fronts. We create a multilingual VQA benchmark in 7 diverse languages to drive modeling progress on multilingual VQA. We establish strong unified and open-ended VQA models that work well on 13 languages as well as benchmark state-of-the-art models. For future work, we would like to expand native-language question generation that is done in a limited scope and have single one for all target answers.

%% file: supp.tex

\begin{figure}[ht!]
\centering
\resizebox{\linewidth}{!}{%
\includegraphics{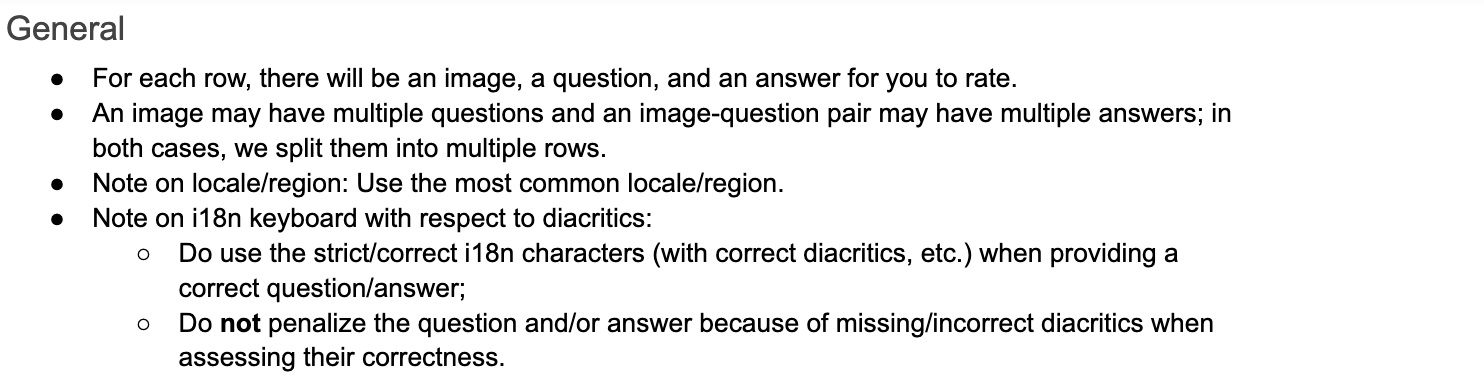}} \\
\resizebox{\linewidth}{!}{%
\includegraphics{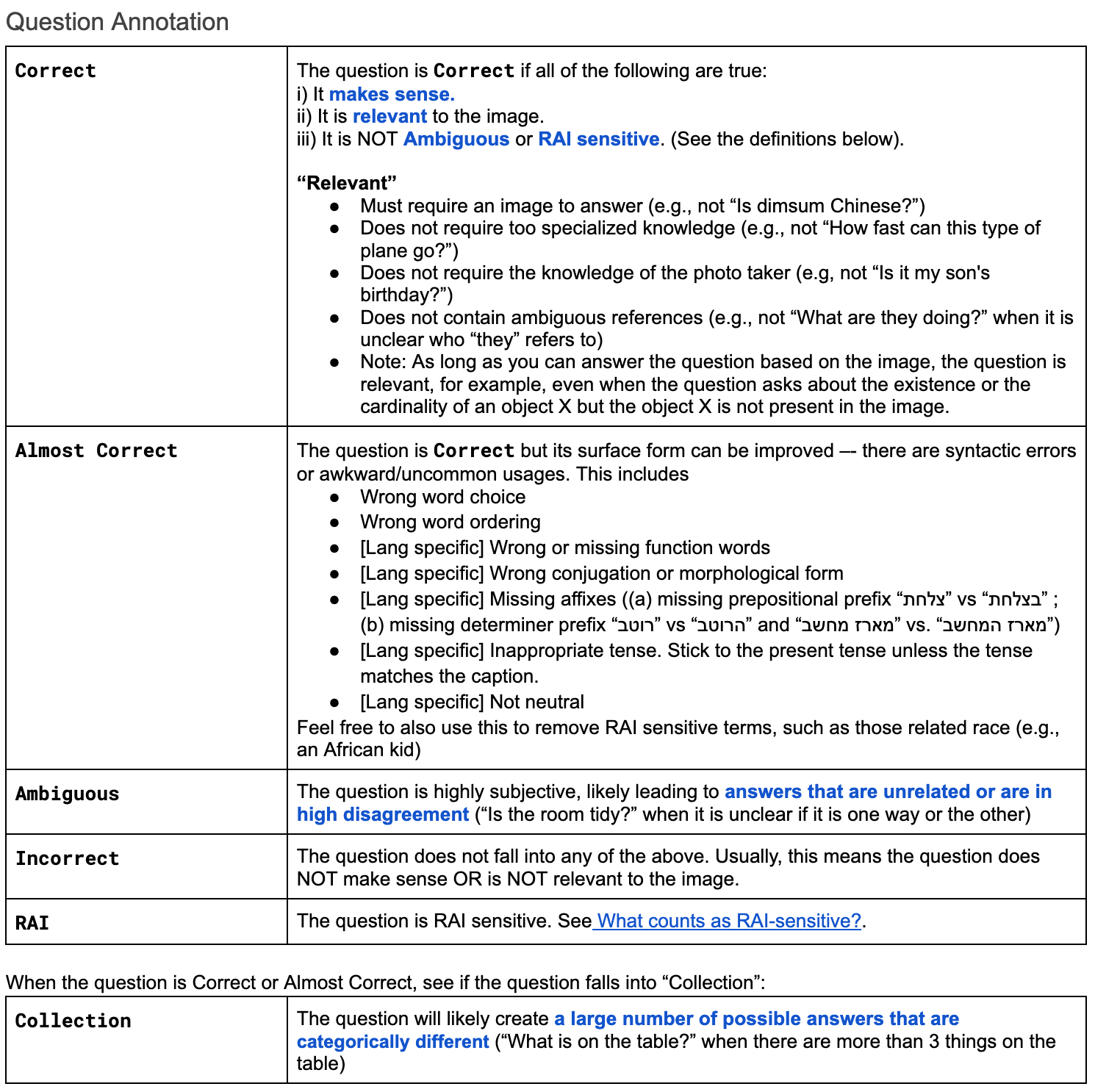}} \\
\resizebox{\linewidth}{!}{%
\includegraphics{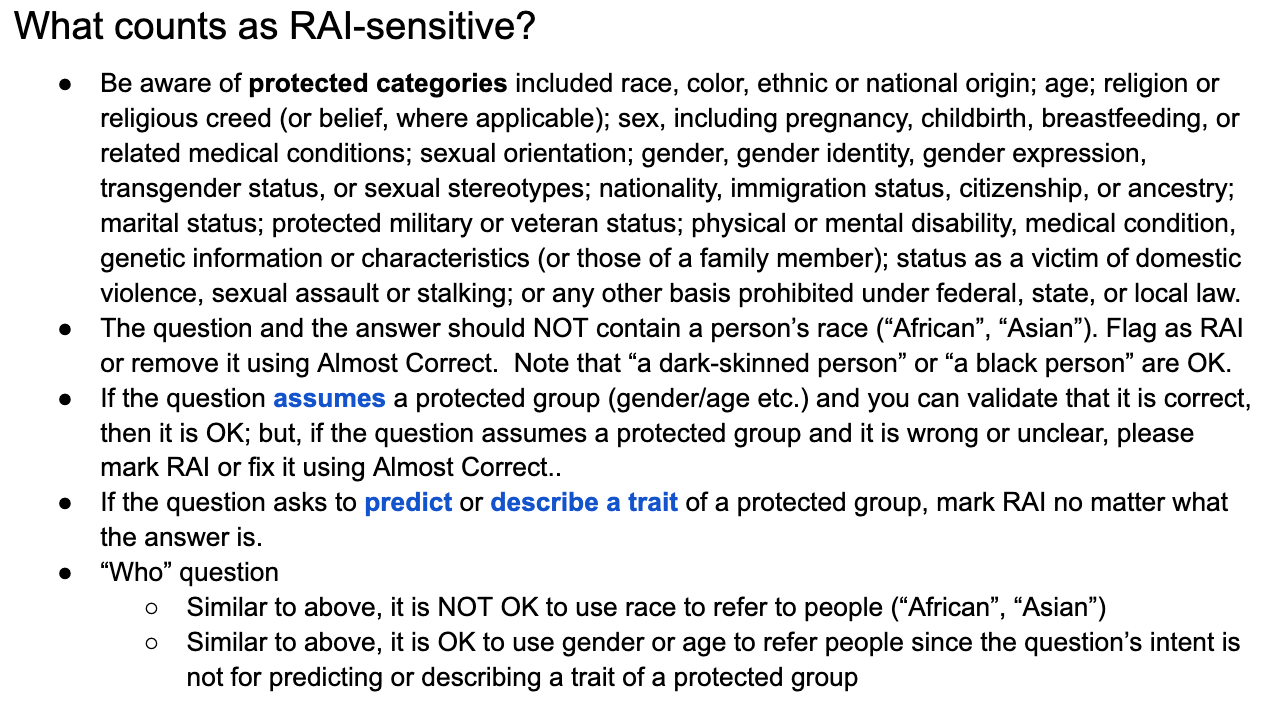}}
\vspace{-15pt}
\caption{\textbf{Detailed Instructions on Question Annotation}}
\label{fig:suppannoq}
\vspace{-10pt}
\end{figure}

\begin{figure}[ht!]
\centering
\resizebox{\linewidth}{!}{%
\includegraphics{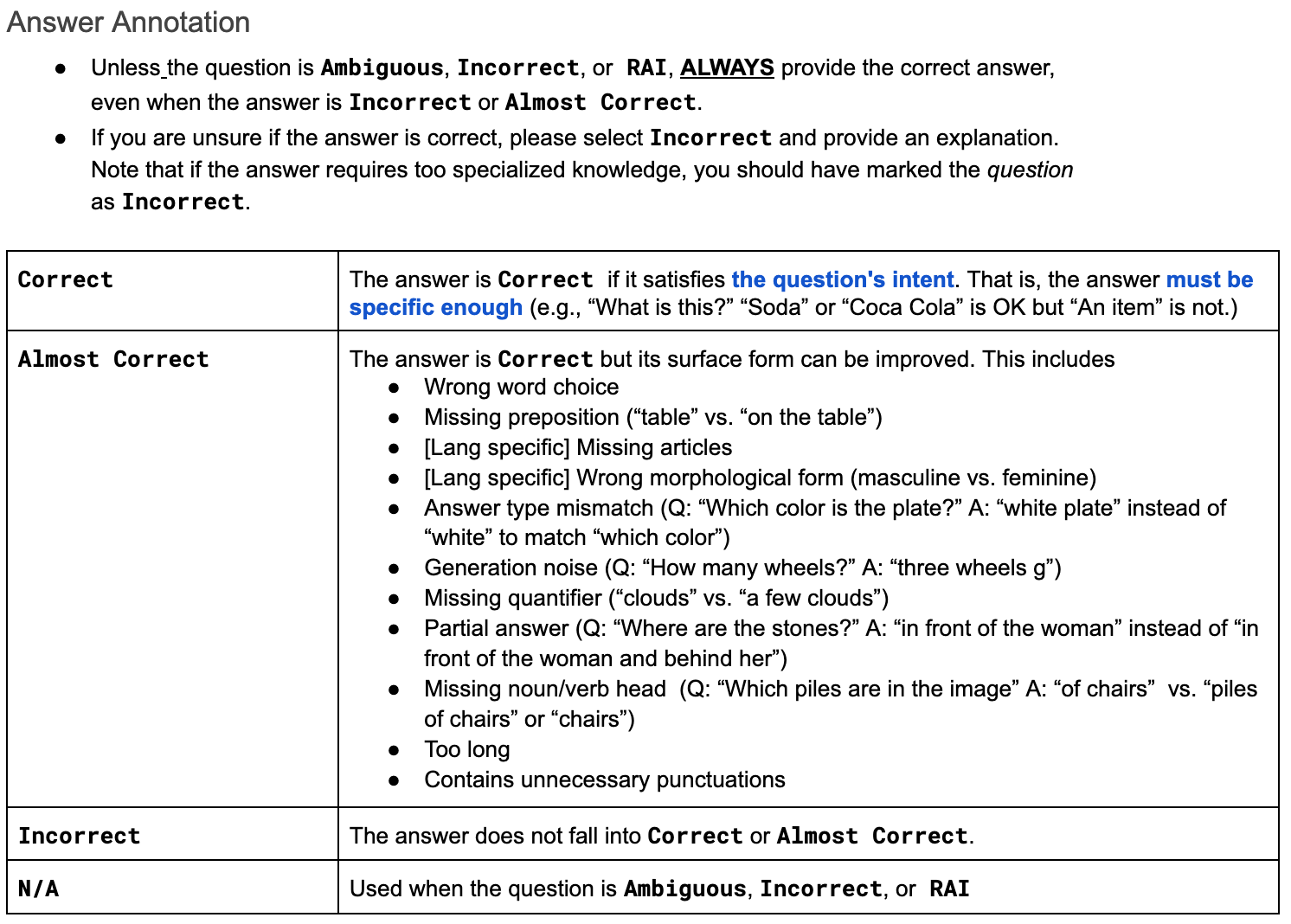}} \\
\resizebox{\linewidth}{!}{%
\includegraphics{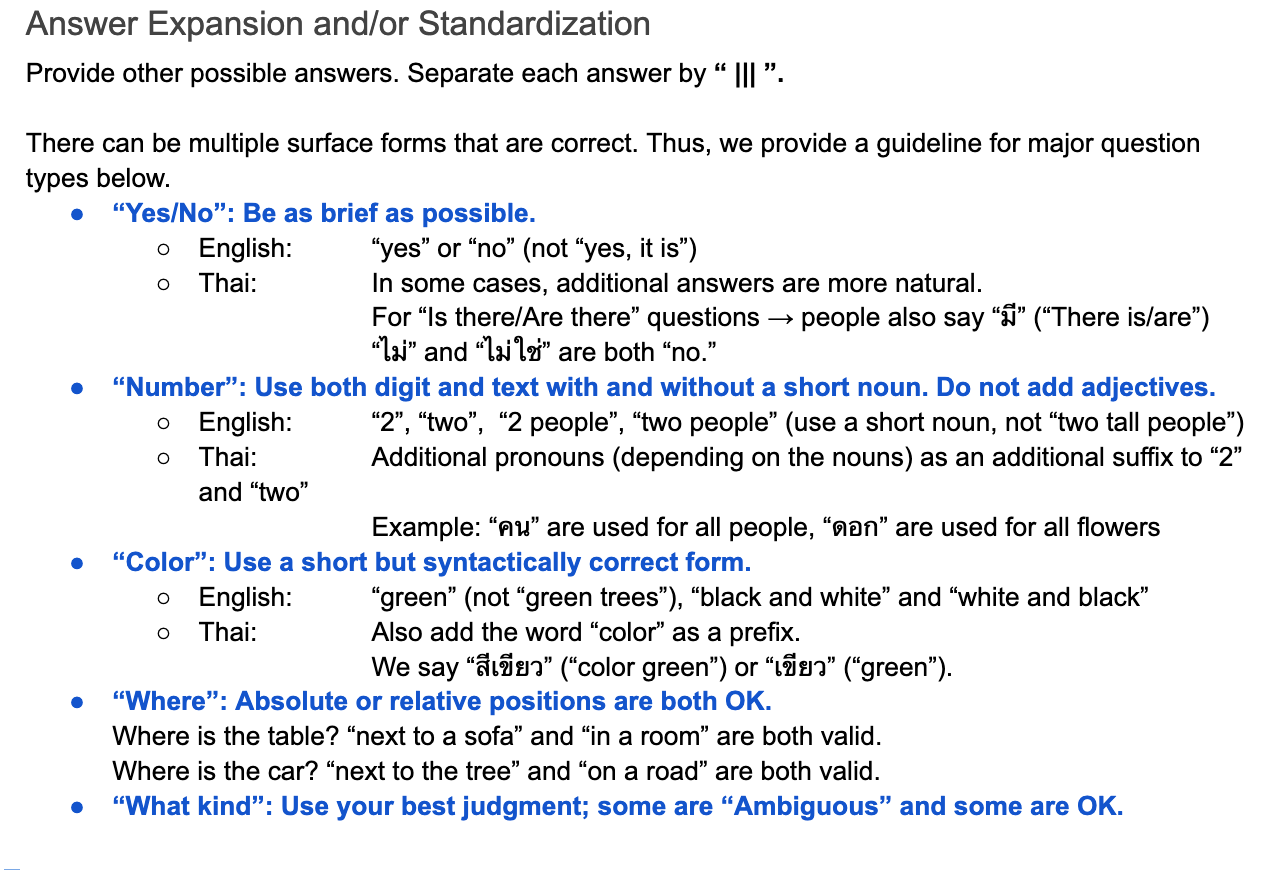}}
\vspace{-15pt}
\caption{\textbf{Detailed Instructions on Answer Annotation 
 as well as Answer Expansion and Standardization.}}
\label{fig:suppannoa}
\vspace{-10pt}
\end{figure}

\begin{figure}[ht!]
\centering
\resizebox{.85\linewidth}{!}{%
\includegraphics{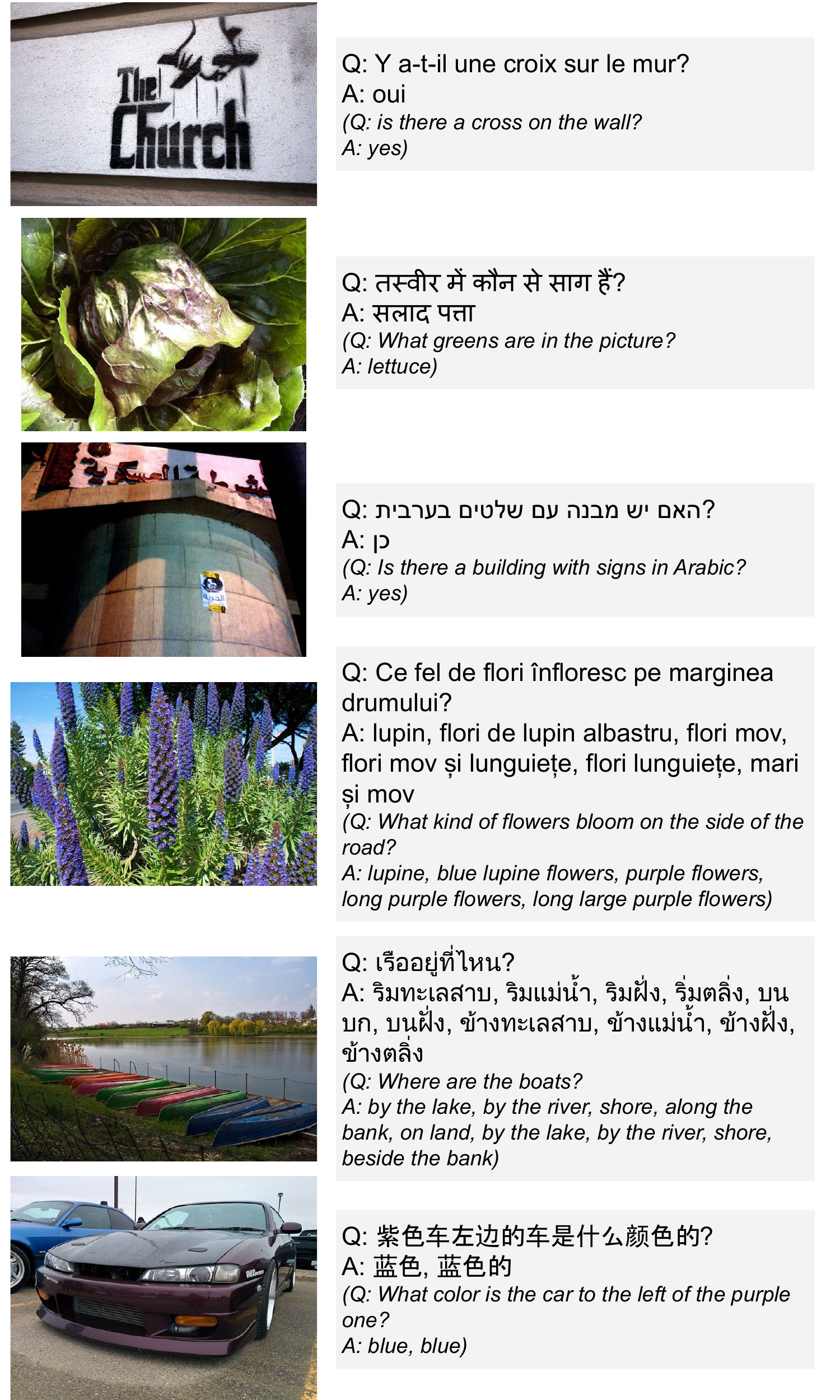}
}
\vspace{-5pt}
\caption{\textbf{Additional MaXM examples.}}
\label{fig:suppintro}
\vspace{-10pt}
\end{figure}

\begin{figure*}[ht!]
\centering
\resizebox{.99\linewidth}{!}{%
\includegraphics{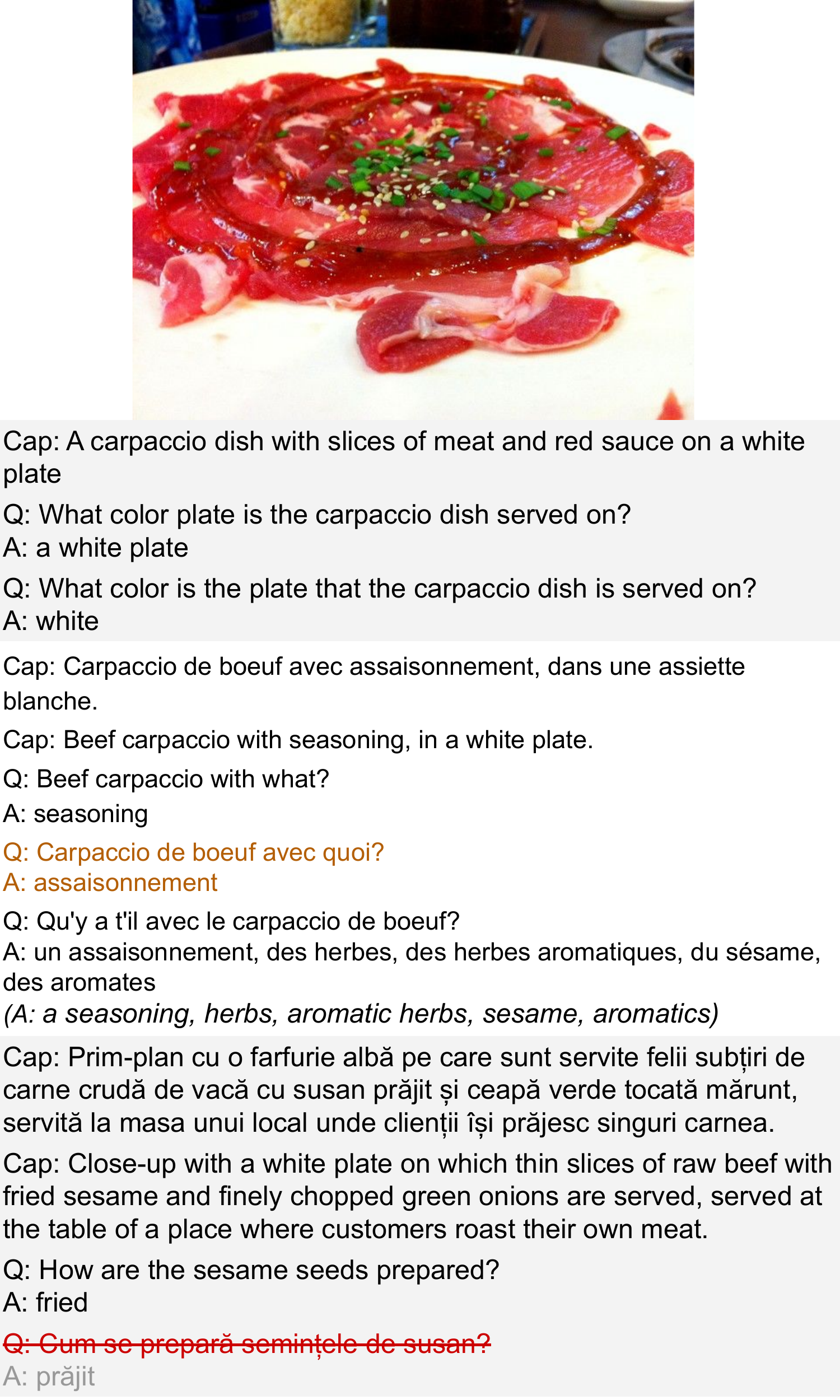}
\includegraphics{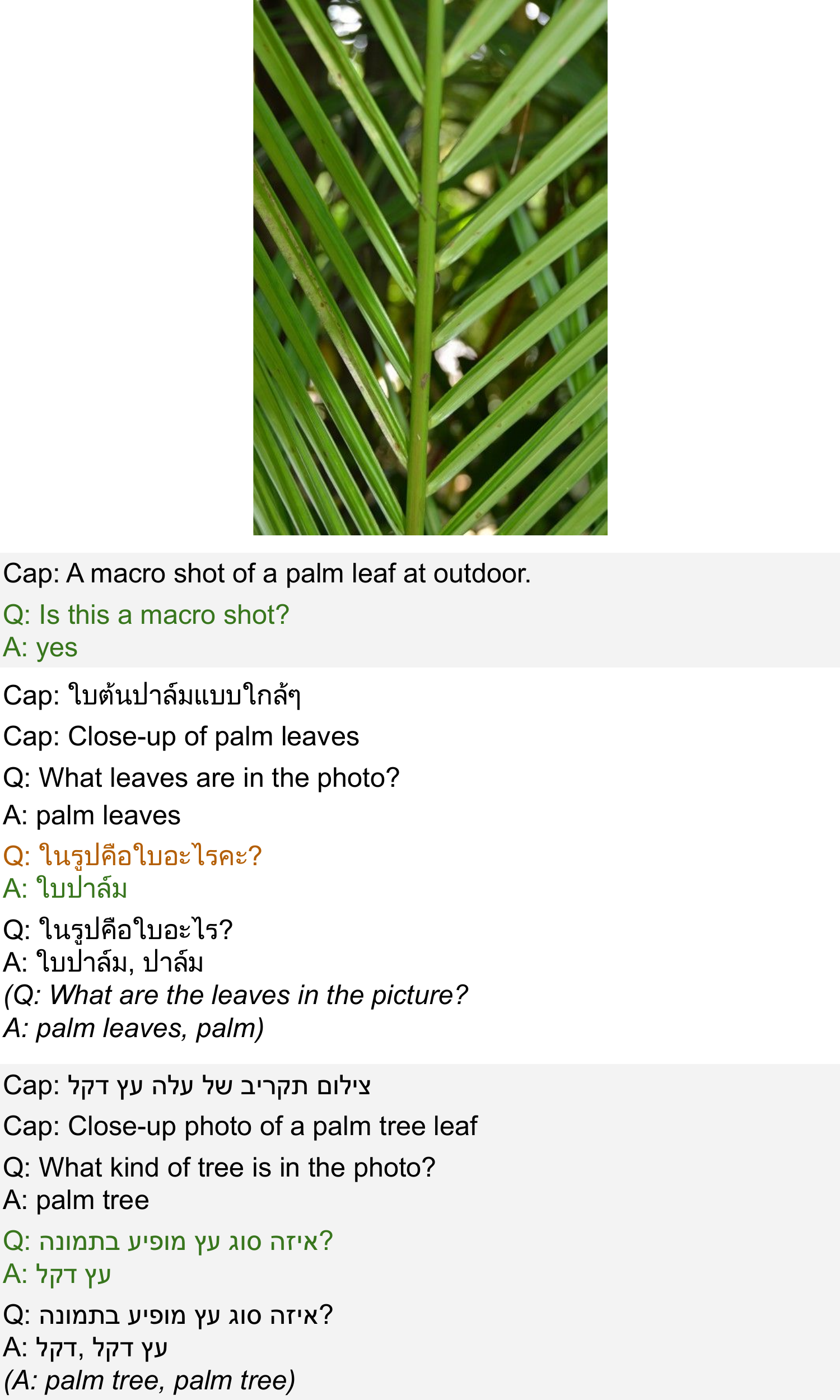}
\includegraphics{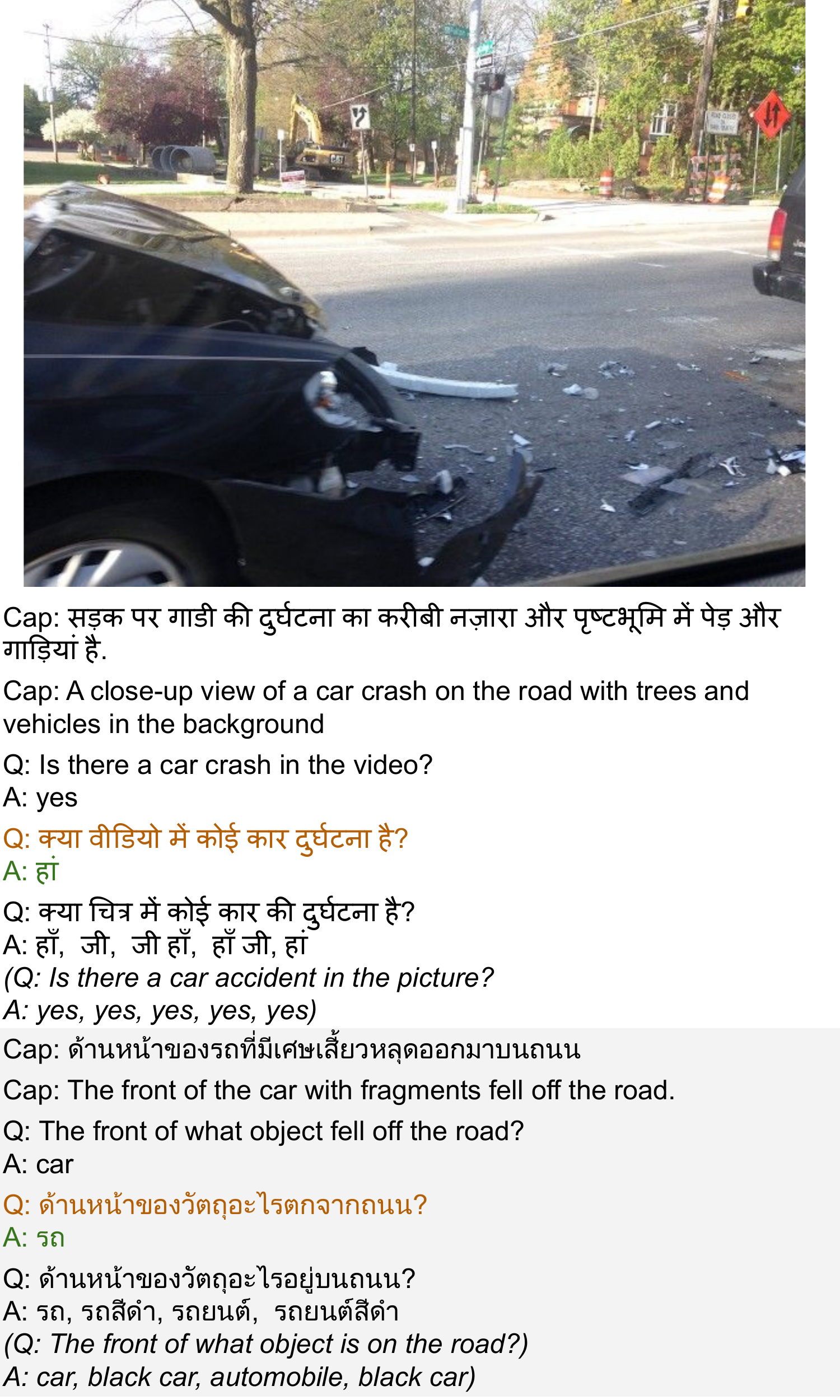}
}
\vspace{-5pt}
\caption{\textbf{Additional examples on our approach to multilingual VQA data generation}. Green, yellow, and red texts correspond to ``Correct'', ``Almost Correct'', and ``Incorrect,'' respectively.}
\label{fig:suppapproach}
\vspace{-10pt}
\end{figure*}

\begin{figure}[ht!]
\centering
\resizebox{.85\linewidth}{!}{%
\includegraphics{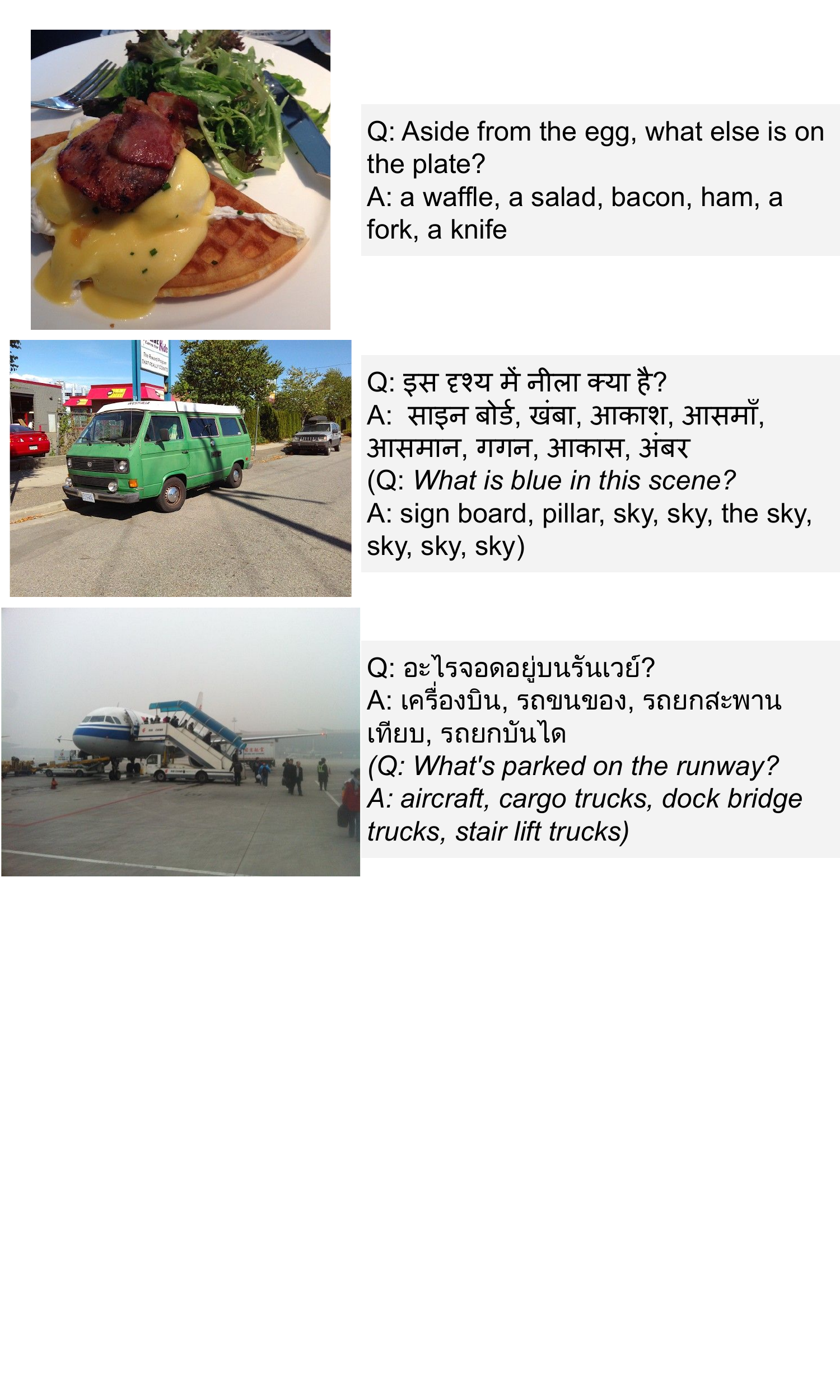}
}
\vspace{-125pt}
\caption{\textbf{Examples of Collection questions.}}
\label{fig:suppcollection}
\vspace{-10pt}
\end{figure}

\begin{figure}[ht!]
\centering
\resizebox{.85\linewidth}{!}{%
\includegraphics{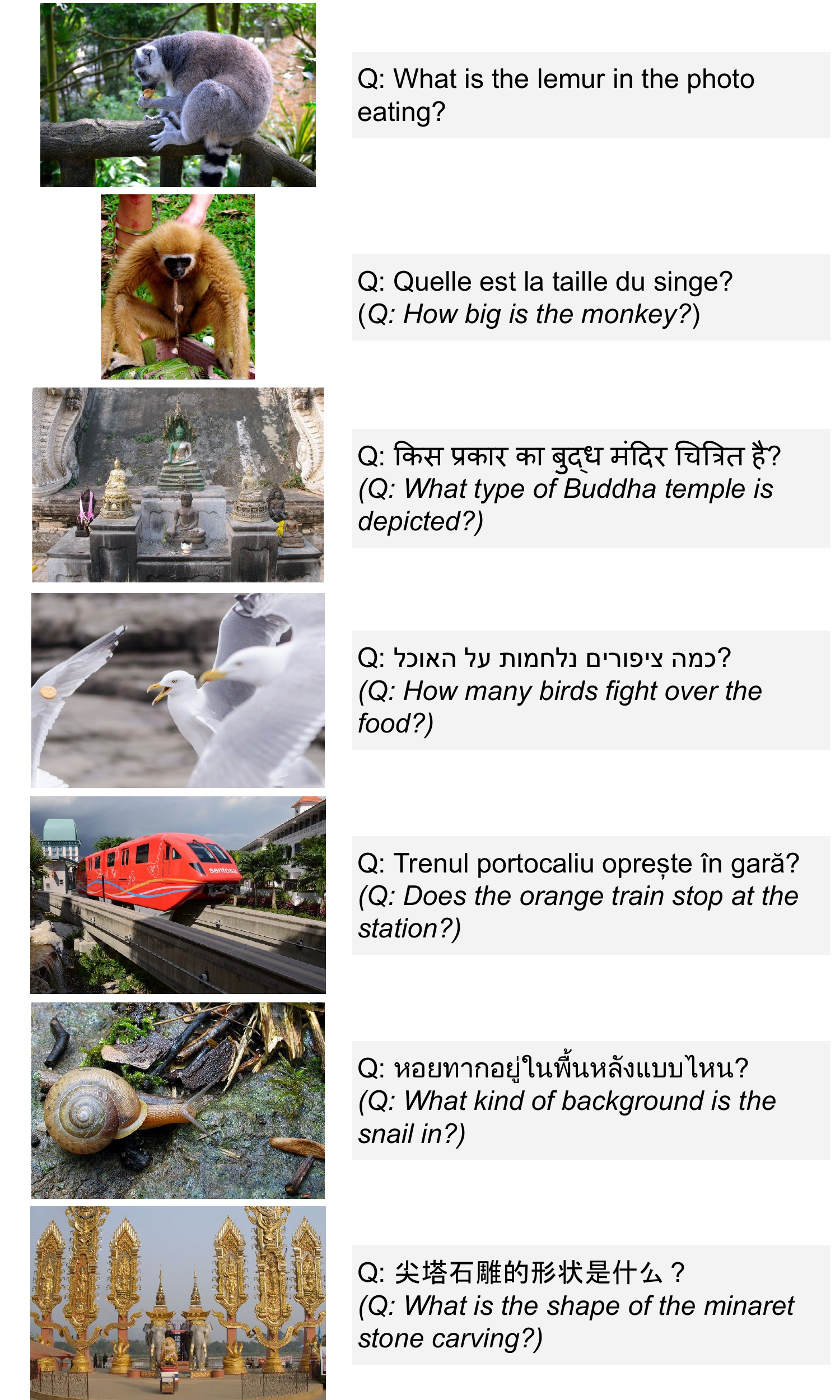}
}
\vspace{-5pt}
\caption{\textbf{Examples of Ambiguous questions} that we flagged and filtered out.}
\label{fig:suppambiguous}
\vspace{-10pt}
\end{figure}

\begin{figure}[ht!]
\centering
\resizebox{.85\linewidth}{!}{%
\includegraphics{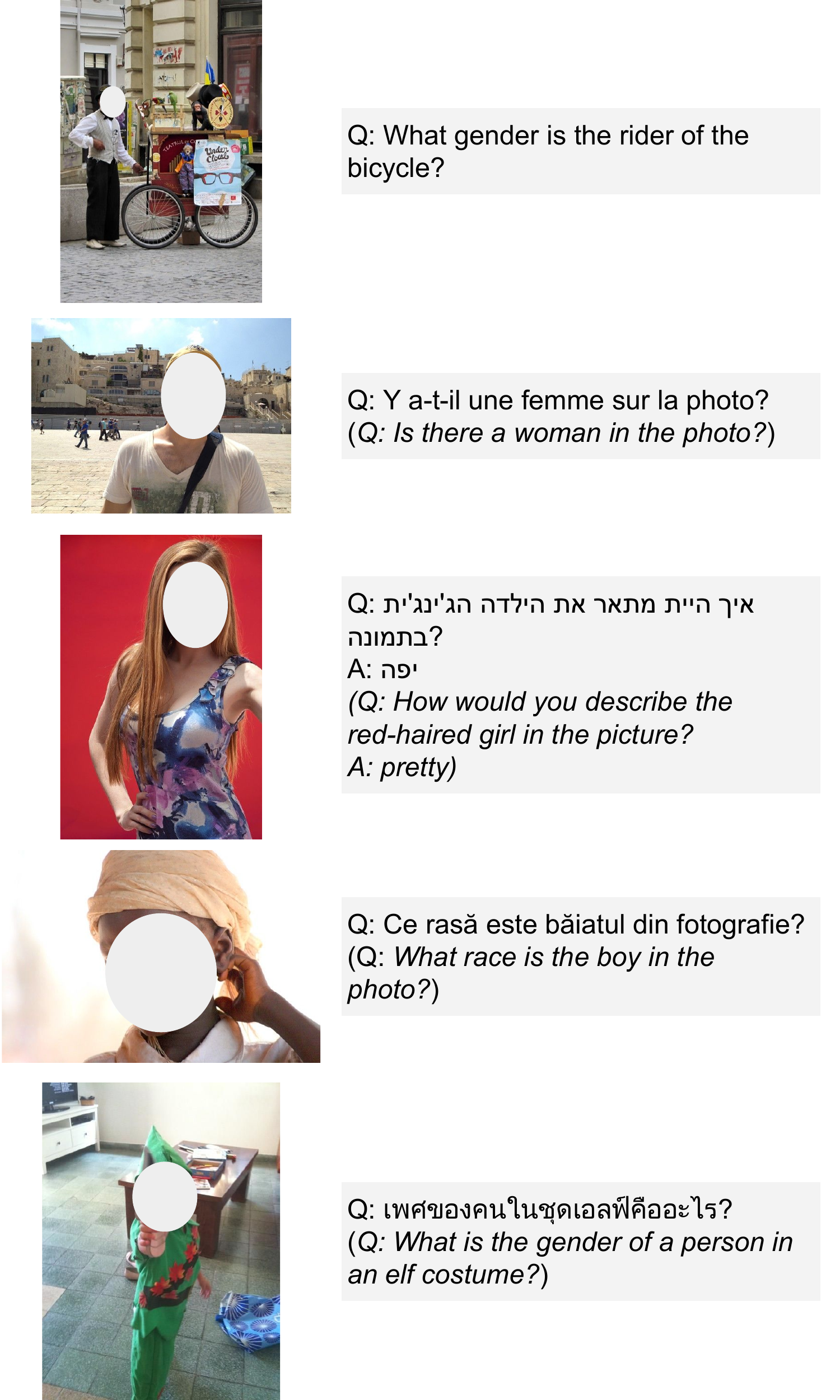}
}
\vspace{-5pt}
\caption{\textbf{Examples of Responsible-AI-sensitive questions} that we flagged and filtered out. Faces are hidden.}
\label{fig:supprai}
\vspace{-10pt}
\end{figure}

\section{Considerations and Limitations}
\label{apdx:consider}

Our dataset is intended to be used for research-only purposes.  

Our pipeline takes in an image caption as input. Image captions may have mistakes and biases, which could be further amplified by machine learning models used by our approach. In particular, we use generative models for automatic question generation and machine translation that may create outputs with incorrect or nonfactual contents or outputs with Translationese artifacts. We have mitigated this manually via human in the loop and automatically via the caption-answer consistency check (cf., Sect.~\ref{ssec:approach_dtvq2a}). Note that the English \qsq{}~\cite{changpinyo2022all} that we leverage in our pipeline also has similar filtering using the round-trip consistency check via question answering. Together these significantly improve the correctness and fluency of our pipeline. In addition, we explicitly mark examples that can be considered Responsible-AI-sensitive, but not necessarily incorrect; see Sect.~\ref{apdx:anno} for details and examples.

Another type of biases is the low coverage of particular types of answers, resulting from the image captions not mentioning the absences of objects or properties. We have also taken a step toward mitigating this. See Sect.~\ref{ssec:approach_dqg}.

Finally, we select a diverse set of languages, alleviating typological, genealogical, and geographical language biases presented in the VQA research community.

We mainly use Crossmodal-3600 (\xmodal{})~\cite{xm3600}. Open Images~\cite{openimages,oidv4} and the multilingual captions in \xmodal{} are human-curated and cleaned, which mitigates the risks that \maxm{} would contain information that names or uniquely identifies individual people or offensive content.

\section{Human Verification and Modification}
\label{apdx:anno}

\subsection{Annotation Guideline}
\label{apdx:annoguide}

We provide our general instructions and detailed instructions on question annotation in Fig.~\ref{fig:suppannoq}, where we explicitly ask the annotators to be wary of Responsible-AI-sensitive questions. Fig.~\ref{fig:suppannoa} provides detailed instructions on
answer annotation and on answer expansion and standardization.

\subsection{Additional Examples}
\label{apdx:addex}

\mypartop{Additional Examples}
Fig.~\ref{fig:suppintro} provides additional examples to the ones in Fig.~\ref{fig:intro}. Again, we highlight the richness and diversity of our questions. For instance, it requires recognizing a cross under occlusion (French), a type of vegetables (Hindi), the Arabic language (Hebrew), and a type of flowers (Romanian). Some of these examples are specific to particular languages; it would be difficult for other language speakers to answer the Hebrew example (or the Chinese example in Fig.~\ref{fig:intro}, which requires OCR).

We also highlight the richness of our candidate answers. For the ``where'' question in Thai, 10 answers count as correct. Similarly, the Romanian example in Fig.~\ref{fig:intro} provides multiple diverse surface forms for ``coffee with cream.''

Fig.~\ref{fig:suppapproach} additional examples to the Chinese one in Fig~\ref{fig:create_data}. These examples showcase the efficiency of our annotation process. They also provide concrete examples of ``Almost Correct.'' For instance, in the middle example, the Thai translation of ``What leaves are in the photo?'' is not \emph{neutral} because it contains an Honorific particle \footnote{\tiny{\url{https://en.wikipedia.org/wiki/Thai_honorifics}}}; it ends with ``kh{{\'a}'' which signifies a sign of respect to the addressee and indicates that the sex of the speaker is female. Finally, these examples provide a glimpse of sources of errors. For instance, it is \qsq{} that hallucinates ``in the video'' in the Hindi example on the right.

\mypartop{Collection Examples}
Fig.~\ref{fig:suppambiguous} provides examples of ``Collection'' questions. We keep these questions as we believe they are useful in practice and as a way to encourage the community to work on better automatic evaluation metrics for this type of questions.

\mypartop{Ambiguous Examples}
Fig.~\ref{fig:suppcollection} provides examples of ``Ambiguous'' questions that we filter out. Reasons include object being too small (English) or irregular (Chinese), determining sizes being subjective (French), and not enough context (Hebrew, Romanian). ``What kind/What type'' questions are particularly difficult to answer and tend to be ambiguous.

\mypartop{Responsible-AI-sensitive Examples}
Fig.~\ref{fig:supprai} provides examples of Responsible-AI-sensitive questions that we filter out. These cases are often associated with directly asking for the information about or describing a particular gender or race, or involving an incorrect assumption about such protected attributes (e.g., girl vs. woman in the Hebrew example).

\begin{figure}[t]
\centering
\resizebox{0.9\linewidth}{!}{%
\includegraphics{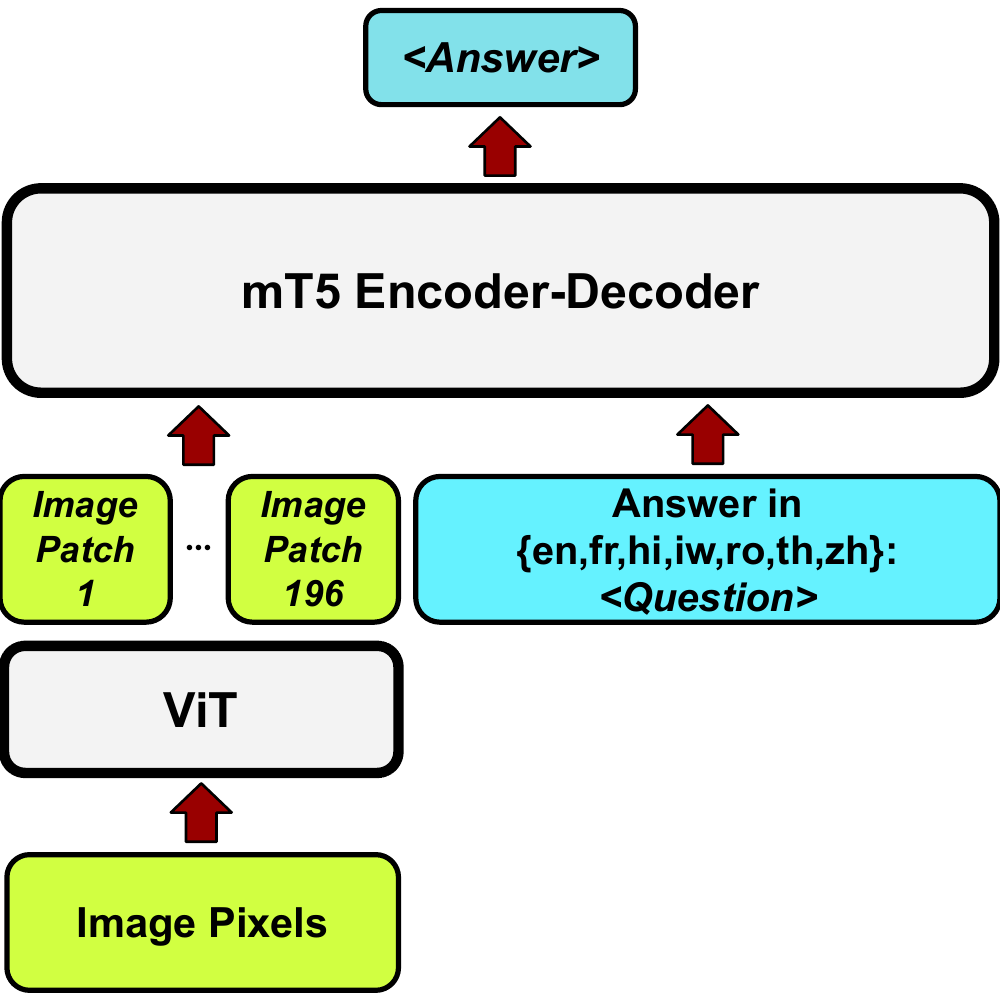}
}
\vspace{-5pt}
\caption{\textbf{Our Simple MPT model} used in our experiments. We leverage ViT~\cite{vit} and mT5~\cite{mt5} and train them together end-to-end.}
\label{fig:model}
\vspace{-15pt}
\end{figure}

\begin{table*}[ht]
\footnotesize
\begin{center}
\begin{tabular}{l|c|c|c|c|c|c|c|c|c@{}}
 & \multicolumn{1}{c|}{Finetuning} & \multicolumn{8}{c}{Question Language} \\  \cline{3-10}
\multicolumn{1}{c|}{Model} & \multicolumn{1}{c|}{Dataset} & en & bn & de & id & ko & pt & ru & zh \\ \hline
M3P~\cite{xgqa} & GQA & \textbf{58.4} & 17.6 & 24.8 & 18.7 & 19.7 & 26.7 & 24.3 & 19.7 \\
mBERT$^{\textrm{Ada}}$~\cite{xgqa} & GQA & \emph{56.3} & 13.4 & 32.4 & 19.8 & 19.9 & 31.5 & 25.5 & 26.2 \\ \hline \hline
Single-Language & \vqaset{}  & 43.1 & 37.9 & 39.6 & \textbf{40.4} & \emph{38.9} & \emph{40.3} & 39.3 & \emph{39.7} \\ \hline
Simple MPT & \vqaset{} & 41.5 & \emph{38.6} & \emph{40.5} & 39.5 & 38.7 & 39.8 & \emph{39.5} & 39.5 \\
Simple MPT & \qsqcoco{} & 36.6 & 34.3 & 36.1 & 35.5 & 35.1 & 34.6 & 34.5 & 35.4 \\
Simple MPT & \qsqcc{} & 34.0 & 30.9 & 33.3 & 33.2 & 32.5 & 32.1 & 32.0 & 32.7 \\ \hline
PaLI~\cite{pali2}& \vqaset{} & 54.2 & \textbf{50.0} & \textbf{52.2} & \textbf{50.6} & \textbf{50.4} & \textbf{51.3} & \textbf{50.3} & \textbf{50.6} \\
\hline
\end{tabular}
\end{center}
\vspace{-5pt}
\caption{\textbf{Zero-Shot Results on xGQA}. Accuracy (\%) of our Simple MPT models trained on different training datasets as well as our Single-Language baselines and the baselines from \cite{xgqa}. Best results are \textbf{bold}. Second best \emph{italicized}.}
\vspace{-5pt}
\label{tab:xgqa_res}
\end{table*}

\section{Simple MPT}
\label{apdx:mpt}

In this section, we describe \textbf{Simple MPT}, a lightweight model for mVQA in detail.

\mypar{Design} Much of the previous work on VQA is built for English. 
Further, VQA is often formulated as \emph{vocab-based VQA}, a classification task into a pre-defined space of top (English) answer vocabulary; see, e.g., \cite{vqa1,vqa2}. The main drawback of this approach is its inability to deal with rare answers through language compositionality. Recent work considers VQA as generation~\cite{cho2021vltt5,wang2022simvlm,flamingo,git}, capable of \emph{open-ended VQA}. We adopt this as a scalable and flexible modeling approach to mVQA as the language coverage increases. In particular, we propose a \emph{single} open-ended VQA model for multiple languages. Our proposed formulation is more desirable than existing ones since it takes advantage of both compositionality in individual languages and the relationship among related languages. To this end, we first describe an encoder-decoder architecture for VQA in the open-ended generation setting. Then, we describe how we train this model for multiple languages. This is summarized in Fig.~\ref{fig:model}.

\mypar{Open-Ended VQA}
Our starting architecture is mT5~\cite{mt5}, a multilingual variant of T5~\cite{t5}. mT5 is an encoder-decoder transformer-based architecture, pre-trained on a Common Crawl-based dataset covering 101 languages. This allows us to leverage multilingual language understanding (for the questions) and generation (for the answers) from the get-go. To adapt mT5 to the VQA task, we prepend patch embeddings from the image to the question tokens. In particular, we encode the image pixels using Vision Transformers (ViT)~\cite{vit}. We use ViT-L16 and mT5-Large in all of our experiments. Both mT5 and ViT are trained together in an end-to-end fashion to predict the target answer for each image-question pair, using the standard cross-entropy loss.

\mypar{Multi-Language Prompted Training}
We resort to multi-task prompted/instruction training ~\cite{sanh2022multitask,wei2022finetuned}, where a task corresponds to VQA for a particular language. For the input question $\langle$question$\rangle$ in language $\langle$lang$\rangle$, we construct the prompt ``Answer in $\langle$lang$\rangle$: $\langle$question$\rangle$'' and use it as the text input to our model, similar to a modification to the input in Google's Multilingual Neural Machine Translation System~\cite{johnson2017google}. Such a design for multi-task learning makes extending VQA to multiple languages simple; as data for additional languages become available, one can simply add them to the pool without the need for architecture changes.

\mypar{Implementation Details} We use the Flax implementation~\cite{jax2018github}. For training both our 2\lang{} and 2en models, we use Adafactor~\cite{adafactor} with a $\beta_1$ of 0 and a second-moment exponential decay of 0.8. We use a linear warmup of 1K steps with a peak learning of learning rate of 1e-3 and inverse square-root decay. We set the ViT dropout rate to 0 and the mT5 dropout rate to 0.1. We train each model with data parallelism using 16 Cloud TPU Pods\footnote{\url{https://cloud.google.com/tpu}}, each with a batch size of 512, for 100K steps. We use standard image resolution of 224x224. We use the maximum input length of 24 and the target output length of 8.

We consider three datasets for training Simple MPT, all are translations of existing large-scale English VQA datasets to the 13 languages covered by \maxm and xGQA. We use the Karpathy training split~\cite{karpathy2015deep} for \vqaset{} and \qsqcoco{} and the standard training split for \qsqcc{}.


\section{Additional Results}
\label{apdx:add_res}

Our Simple MPT in the main paper predicts the answer in the same language as the question. Here, we explore if our Simple MPT can also be useful for the cross-lingual setting in xGQA~\cite{xgqa}, where the model always predicts the answer in English.

Similar to \maxm{}, xGQA is a test-only benchmark, the testdev split of 12,578 question-answer pairs per language from 398 images in 8 languages (en,bn,de,id,ko,pt,ru,zh). To evaluate Simple MPT on this dataset, we use the setting in the main paper: training on \vqaset{}, \qsqcoco{}, and \qsqcoco{} but do not translate the training answers. We also use the prompt ``Answer in en: $\langle$question$\rangle$'' instead of ``Answer in $\langle$lang$\rangle$: $\langle$question$\rangle$''.

Table~\ref{tab:xgqa_res} reports the results. Our baselines are M3P and mBERT$^{\textrm{Ada}}$ from~\cite{xgqa}. Note that both M3P and mBERT$^{\textrm{Ada}}$ have access to the (English) GQA training data~\cite{gqa}, where our model does not. On the other hand, they do not use translated data as in our case.
We outperform \emph{multilingual} zero-shot baselines on all non-English languages, without access to English GQA labeled data. This further confirms that our unified approach to mVQA is effective. In addition, unlike on \maxm{}, \vqaset{} is the best pre-training data source. We attribute this to the fact that \vqaset{} and xGQA share COCO images~\cite{coco}. This highlights the utility of \maxm{} as additional out-of-domain test-only VQA evaluation data.